\documentclass[twoside, accepted]{article}

\usepackage{aistats2025}
\usepackage[round,authoryear]{natbib}

\usepackage[utf8]{inputenc} % allow utf-8 input
\usepackage[T1]{fontenc}    % use 8-bit T1 fonts
\usepackage{hyperref}
\hypersetup{colorlinks=false, hidelinks}
\usepackage{xcolor}
\usepackage{algorithm}
\usepackage[noend]{algorithmic}

\usepackage{amsmath,amssymb,amsthm,mathtools}
\usepackage{footnote}
\usepackage{microtype}      % microtypography

\usepackage{booktabs, arydshln}
\usepackage{multirow}
\usepackage{mdframed}

\newcommand{\bbB}[0]{\mathbb{B}}
\newcommand{\bbE}[0]{\mathbb{E}}
\newcommand{\bbF}[0]{\mathbb{F}}

\newcommand{\bbP}[0]{\mathbb{P}}
\newcommand{\bbQ}[0]{\mathbb{Q}}
\newcommand{\bbR}[0]{\mathbb{R}}

\newcommand{\calC}[0]{\mathcal{C}}
\newcommand{\calL}[0]{\mathcal{L}}
\newcommand{\calM}[0]{\mathcal{M}}

\newcommand{\calP}[0]{\mathcal{P}}
\newcommand{\calR}[0]{\mathcal{R}}

\DeclareMathOperator*{\argmin}{arg\,min}
\newcommand{\dd}{\mathop{}\!\mathrm{d}} % differential operator
\newcommand{\kl}[2]{\operatorname{KL}\left(#1 \mid\mid #2\right)}  % Kullback-Leibler

\newcommand{\norm}[1]{\left\lVert #1 \right\rVert}
\newcommand{\cond}{{\;|\;}}

\newcommand{\mrm}[1]{\mathrm{#1}}

\makeatletter
\def\adl@drawiv#1#2#3{%
	\hskip-.314159\tabcolsep
	\xleaders#3{#2.5\@tempdimb #1{1}#2.5\@tempdimb}%
	#2\z@ plus1fil minus1fil\relax
	\hskip-.314159\tabcolsep}
\newcommand{\cdashlinelr}[1]{%
	\noalign{\vskip\aboverulesep
		\global\let\@dashdrawstore\adl@draw
		\global\let\adl@draw\adl@drawiv}
	\cdashline{#1}
	\noalign{\global\let\adl@draw\@dashdrawstore
		\vskip\belowrulesep}}
\makeatother

\newcommand{\codeaddress}{https://github.com/zgbkdlm/fbs}

\usepackage[shortlabels]{enumitem}
\setlist[enumerate]{itemsep=1pt, topsep=1pt, leftmargin=15pt}
\setlist[itemize]{itemsep=1pt, topsep=1pt}

\mdfsetup{skipbelow=-1pt}

\mdfdefinestyle{enumFrame}{%
    linecolor=gray!50!white,
    outerlinewidth=2pt,
    innertopmargin=4pt,
    innerbottommargin=4pt,
    innerrightmargin=4pt,
    innerleftmargin=1pt,
    leftmargin = 1pt,
    rightmargin = 4pt,
    backgroundcolor=gray!10!white,
    nobreak=true,
}

\theoremstyle{plain}
\newtheorem{theorem}{Theorem}[section]

\theoremstyle{definition}

\theoremstyle{remark}
\newtheorem{remark}[theorem]{Remark}

\begin{document}



\twocolumn[

\aistatstitle{Conditioning diffusion models by explicit forward-backward bridging}

\aistatsauthor{Adrien Corenflos$^{\textbf{*}1}$ $\quad$ Zheng Zhao$^{\textbf{*}2}$ $\quad$ Simo S\"{a}rkk\"{a}$^3$ $\quad$ Jens Sj\"{o}lund$^4$ $\quad$ Thomas B. Sch\"{o}n$^4$}

\aistatsaddress{ University of Warwick$^1$ $\quad$ Link\"{o}ping University$^2$ $\quad$ Aalto University$^3$ $\quad$ Uppsala University$^4$}]

\begin{abstract}
    Given an unconditional diffusion model targeting a joint model $\pi(x, y)$, using it to perform conditional simulation $\pi(x \mid y)$ is still largely an open question and is typically achieved by learning conditional drifts to the denoising SDE after the fact.
    In this work, we express \emph{exact} conditional simulation within the \emph{approximate} diffusion model as an inference problem on an augmented space corresponding to a partial SDE bridge.
    This perspective allows us to implement efficient and principled particle Gibbs and pseudo-marginal samplers marginally targeting the conditional distribution $\pi(x \mid y)$.
    Contrary to existing methodology, our methods do not introduce any additional approximation to the unconditional diffusion model aside from the Monte Carlo error.
    We showcase the benefits and drawbacks of our approach on a series of synthetic and real data examples.
\end{abstract}

\section{INTRODUCTION}\label{sec:introduction}
Denoising diffusion models~\citep{song2021scorebased, Ho2020} have recently received a lot of attention as general-purpose samplers for generative modelling in diverse fields such as image analysis~\citep{Luo2023IRSDE}, protein folding~\citep{wu2023practical}, and statistical inference~\citep{Vargas2023denoising}.
At the core, given a target distribution $\pi_0$ that we want to generate samples from, and a time-homogeneous stochastic differential equation~\citep[SDE, see, e.g.,][]{Karatzas1991}
\begin{equation}
    \label{eq:fwd-sde-base}
    \dd X_t = f(X_t) \dd t + \dd W_t, \qquad  X_0 \sim \pi_0,
\end{equation}
with stationary distribution $\pi_{\mrm{ref}}$, they sample from $\pi_0$ by ``denoising''~\eqref{eq:fwd-sde-base} via Doob's $h$-transform~\citep{Rogers2000diffusion}.
Formally, let $\pi_t$ be the distribution of $X_t$ under~\eqref{eq:fwd-sde-base}, and assume that we know how to sample from $\pi_T$ for some $T > 0$.
We can obtain approximate samples from $\pi_0$ by sampling from
\begin{equation}
    \label{eq:bwd-sde-base}
    \begin{split}
        \dd U_t = f_{\mrm{rev}}(U_t, t) \dd t + \dd B_t, \quad U_0 \sim \pi_T,
    \end{split}
\end{equation}
where $B$ is another Brownian motion, and we write $f_{\mrm{rev}}(U_t, t) \coloneqq -f(U_t)  + \nabla \log \pi_{T-t}(U_t)$.
When $T \gg 1$, and under ergodicity guarantees~\citep{Meyn2009}, $\pi_T \approx \pi_{\mrm{ref}}$.
Consequently, if~\eqref{eq:fwd-sde-base} is chosen such that $\pi_{\mrm{ref}}$ is easy to sample from, then the only remaining difficulty is computing the score $\nabla \log \pi_{T-t}$, which is typically overcome by training a neural network to approximate the score function (or drift) via score matching~\citep{hyvarinen2005score}.

Once the score function has been computed, approximate samples from $\pi$ can be obtained by running a discretised simulation of~\eqref{eq:bwd-sde-base} as
\begin{equation}
    \label{eq:score-loss-discrete}
    \begin{split}
        U_{k} &= U_{k-1} - \Delta_k \, f_{\mrm{rev}}(U_{k-1}, t_{k-1}) + \sqrt{\Delta_k} \, Z_k,\\
        U_{0} &\sim \pi_{\mrm{ref}}, \quad Z_k \sim \mathcal{N}(0, I_d),
    \end{split}
\end{equation}
at any grid $0 = t_0 < \cdots < t_K = T$, where $I_d$ is a $d$-dimensional diagonal matrix, and $\Delta_k=t_k - t_{k-1}$.

A typical choice for the forward SDE~\eqref{eq:fwd-sde-base} is a Langevin diffusion targeting a standard Gaussian distribution, namely taking $f(x) = -x$, also known as the Ornstein--Uhlenbeck process.
Variants of this choice have been proposed, such as the use of time-dependent diffusion coefficient~\citep{Song2020improve} or non-Markovian dynamics~\citep{song2020denoising}.
A remarkable extension is the use of Schr\"{o}dinger bridges~\citep{deBortoli2021diffusion}, which generalise the approach by forming SDE bridges between two arbitrary distributions, thereby allowing for the generation of samples from $\pi_0$ given samples from $\pi_T = \pi_{\mrm{ref}}$, in finite time rather than asymptotically as $T \gg 1$ at the cost of learning both the forward and backward dynamics.

    \subsection{PROBLEM FORMULATION}\label{sec:problem-formulation}
    Despite the success of diffusion models, performing \emph{exact} conditional simulation within these is still largely an open question, whereby the goal is to sample from the conditional distribution $\pi(x \mid y)$ given a diffusion model representing the joint distribution~$\pi(x, y)$.
    In this article, we assume that the model for $\pi(x, y)$ is correct and can be trusted so that we typically ignore unconditional modelling errors. 
    We therefore consider the problem of exact conditioning \emph{within} diffusion models. 
    This is a question posed by Bayesian inference in general~\citep{martin2023computing}, also known as \emph{inpainting} and \emph{super-resolution} within the computer vision community.
    
    We assume that a diffusion model (e.g., Schr\"{o}dinger bridge) has been trained to sample from the joint distribution $\pi(x, y)$, and our aim is to produce (asymptotically) exact samples from the conditional distribution $\pi(x \mid y)$ on top of the trained model.

    \subsection{CONTRIBUTIONS}
    We propose a new Markov chain Monte Carlo~\citep[MCMC,][]{Meyn2009} method for conditioning \emph{within} diffusion models, to obtain samples from $\pi(x \mid y)$.
    Our method, which we call forward-backward bridging (FBB), is based on the following observation: conditioning a diffusion model can be written as a partial SDE bridge, constructed by running a preliminary forward simulation, noising the initial states $(X_0, y)$ into $(X_T, Y_T)$, and then a backward particle filter corresponding to bridging back to the target distribution $\pi_0(x \mid y)$.
    This perspective was originally proposed in~\citet{trippe2023diffusion} and in the work, concurrent to ours, of~\citet{dou2024diffusion}

    In contrast to both these works, which approximate the conditional sampling procedure by ignoring errors arising from the use of a backward particle filter, here we take a more principled approach of treating the conditioning problem as an inference problem on the joint $p(x_0, x_T, y_T \mid Y_0=y)$, as well as all intermediary missing steps.
    We achieve this by leveraging the dynamic structure of diffusion models and alternating between a forward noising process and a conditional sequential Monte Carlo~\citep[CSMC,][]{andrieu2010particle} algorithm, thereby forming a Gibbs sampler~\citep{Geman1984stochasticrelaxation} alternatively sampling from the joint forward and conditional backward distributions of the diffusion model.
    This presents the substantial benefits of correcting the sampler for lack of ergodicity~\citep[i.e., the fact of using finite number of particles, and that $T < \infty$ being finite create biases in][]{trippe2023diffusion,dou2024diffusion} as well as providing a way to treat Schr\"odinger bridge  samplers~\citep{deBortoli2021diffusion,shi2023diffusion} which cannot be handled by existing methodology.

    The contributions of this article are therefore as follows:
    \begin{enumerate}
        \item We develop forward-backward bridging (FBB), a new particle Gibbs~\citep{andrieu2010particle} method for conditioning diffusion models. 
        \item In Section~\ref{sec:separable}, we show how, when the noising process is tractable and separable, which is the case for the usual denoising diffusion models, the method can be adapted to be run in-place, at zero memory cost using a pseudo-marginal approach~\citep{andrieu2010particle}. 
        This provides a principled and unbiased generalisation of~\citet{trippe2023diffusion}.
        \item The benefits of our approach are illustrated in Section~\ref{sec:experiments} on a series of high-dimensional benchmarks, showcasing the improved sample quality.
    \end{enumerate}

    \section{FORWARD-BACKWARD CONDITIONING OF DIFFUSIONS}\label{sec:fwd-bwd-conditioning}
    Let us consider a target density $\pi(x, y) = \pi(y) \, \pi(x \mid y)$ for which we have a diffusion sampler, and a value $y$ for conditioning. 
We remark that in contrast to~\citet{Shi2022conditional,wu2023practical,song2021scorebased}, we \emph{do not} assume the capability to sample from or evaluate $\pi(y \mid x)$.
This section describes how we sample from $\pi(x \mid y)$ given a diffusion sampler for $\pi(x, y)$ without any additional training.

\subsection{AN ABSTRACT GIBBS SAMPLER FOR THE CONDITIONAL DISTRIBUTION}\label{subsec:gibbs-sampler}
%
% Let us assume that we have defined a forward ``noising'' diffusion model
Let us define a forward ``noising'' diffusion model
\begin{equation}
    \label{eq:forward-diffusion}
    \begin{split}
        \dd Y_t &= \mu^Y(X_t, Y_t, t) \dd t + \sigma_t \dd W^Y_t, \\
        \dd X_t &= \mu^X(X_t, Y_t, t) \dd t + \sigma_t \dd W^X_t,
    \end{split}
\end{equation}
with $(X_0, Y_0) \sim \pi$ and such that the pair $(X_T, Y_T)$ is distributed according to \emph{any} reference measure $\pi_T = \pi_{\mrm{ref}}$.
We denote the measure associated with the diffusion~\eqref{eq:forward-diffusion} by $\bbF$, and we simplify using the same dispersion $\sigma_t$ although we are not limited to this setting.
This type of SDE is typically given by standard diffusion models~\citep{song2021scorebased} or Schrödinger bridges~\citep[][see also Appendix~\ref{app:background}]{deBortoli2021diffusion}.

The forward diffusion model can be associated with a denoising reverse-time diffusion model
\begin{equation}
    \label{eq:denoising-diffusion}
    \begin{split}
        \dd V_t &= \mu^V(U_t, V_t, t) \dd t + \sigma_t \dd W^V_t, \\
        \dd U_t &= \mu^U(U_t, V_t, t) \dd t + \sigma_t \dd W^U_t,
    \end{split}
\end{equation}
such that if $(U_0, V_0) \sim \pi_{\mrm{ref}}$ then $(U_T, V_T) \sim \pi$ is distributed according to the target measure $\pi(x, y)$.
We denote the measure associated with the backward diffusion~\eqref{eq:denoising-diffusion} by $\bbB$.
Our conditional sampler construction is based on the following abstract Gibbs sampler~\citep{Geman1984stochasticrelaxation}, which corresponds to alternatively propagating~\eqref{eq:forward-diffusion} from $t=0$ to $t=T$, and then sampling $U_T$ conditionally on $U_0=X_T$, $V_0=Y_T$, and $V_T = y$, coming from the forwards simulation pass.
Formally, suppose that, at a Gibbs step $j \geq 0$, the path sample $X_{0}^j$ is distributed according to $\pi(x \mid y)$ for a given $y$, and consider the following procedure:
\begin{mdframed}[style=enumFrame]
\begin{enumerate}
    \item sample $\bigl( Y^{j+1}_{(0, T]}, X_T^{j+1} \bigr)\sim \bbF(\cdot \mid X^j_{0}, Y_0=y)$; \label{item:sample-forward-path-full}
    % \item sample $X^{j+1}_0 = U^{j+1}_{T} \sim \bbB^U(\cdot \mid V^{j+1}_{[0, T]})$. \label{item:sample-backward-path-full}
    \item sample $X^{j+1}_0 = U^{j+1}_{T} \sim \bbB(\cdot \mid U_0 = X_T^{j+1}, V_{[0, T]} = Y^{j+1}_{[T,0]})$. \label{item:sample-backward-path-full}
\end{enumerate}
\end{mdframed}

Then, $X^{j+1}_{0}$ distributes according to the target $\pi(x \mid y)$.

In general, it is hard to directly apply the steps~\ref{item:sample-forward-path-full} and~\ref{item:sample-backward-path-full} above.
Indeed, (i) at evaluation time, diffusion models are given in a discretised form, not in their continuous-time formulation, and (ii), even in their discretised form, the distributions arising in Steps~\ref{item:sample-forward-path-full} and~\ref{item:sample-backward-path-full} are not available in closed form.
Next, we describe a practical MCMC algorithm targeting the discretised version of the two steps.

\subsection{DISCRETE-TIME FORMULATION}\label{subsec:discrete-time-formulation}
In Section~\ref{subsec:gibbs-sampler}, we have described an abstract procedure for sampling from the conditional model $\pi(\cdot \mid y)$ by iterative sampling from a conditioned forward noising process (step~\ref{item:sample-forward-path-full}) and then running a conditioned version of the denoising SDE (step~\ref{item:sample-backward-path-full}).

For this procedure to be practically implementable, we now assume that we have access to the discretised forward and backward SDEs, given by
\begin{equation}
    \label{eq:discrete-model}
    \begin{split}
        Y_{t_{k+1}} \sim \bbF^Y_{t_{k}}(\cdot \mid Y_{t_k}, X_{t_k}), \\
        X_{t_{k+1}} \sim \bbF^X_{t_{k}}(\cdot \mid Y_{t_k}, X_{t_k}),
    \end{split}
    \quad
    \begin{split}
        V_{t'_{k+1}} \sim \bbB^V_{t'_{k}}(\cdot \mid V_{t'_k}, U_{t'_k}), \\
        U_{t'_{k+1}} \sim \bbB^U_{t'_{k}}(\cdot \mid V_{t'_k}, U_{t'_k}),
    \end{split}
\end{equation}
where $t'_k = T- t_k$, $k=0, \ldots, K$,
and form the diffusion model at hand\footnote{For simplicity, we use the same notation for the continuous-time SDE and its (approximate) discretisation.}.

Writing hereafter $X_k$ and $Y_k$ for $X_{t_k}$ and $Y_{t_k}$ (and similarly for $U$, $V$, $\bbF$, and $\bbB$), as well as the shorthands $X_{0:K} = (X_{k})_{k=0}^K$ and $X_{K:0} = (X_{k})_{k=K}^0$ for $K$ time steps, the procedure described in Section~\ref{subsec:gibbs-sampler} then becomes:
\begin{mdframed}[style=enumFrame]
\begin{enumerate}
    \item sample $\bigl( Y^{j+1}_{1:K}, X^{j+1}_{K} \bigr) \sim \bbF(\cdot \mid x_0=X^j_{0}, y_0=y)$; \label{step:forward-discrete-filtering} 
    \item sample $X^{j+1}_{0} = U_{K}^{j+1} \sim \bbB(\cdot \mid v_{0:K} = Y^{j+1}_{K:0}, u_0=X^{j+1}_K)$. \label{step:backward-discrete-filtering}
\end{enumerate}
\end{mdframed}

This structure of the conditional sampler was first introduced in~\citet[albeit not explicitly as a Gibbs sampler]{trippe2023diffusion} for the special case when the drift in~\eqref{eq:fwd-sde-base} is separable, namely when the drifts $\mu^X$ and $\mu^Y$ in \eqref{eq:fwd-sde-base} do not depend on both $Y$ and $X$, respectively (we come back to this special case in Section~\ref{sec:separable}).

The simulation in step~\ref{step:forward-discrete-filtering} is directly implementable.
On the other hand, step~\ref{step:backward-discrete-filtering} a well-studied problem in the literature on Bayesian filtering~\citep[see, e.g.,][]{sarkka2023bayesian}, and approximate solutions are available, for example, in the form of particle filters~\citep[for a recent review of these see][]{chopin2020book} which were employed in~\citet{trippe2023diffusion}.
However, for a finite number of Monte Carlo samples used, these methods provide biased (despite consistency) samples from the filtering distribution, therefore resulting in a biased sampler for targeting $\pi(\cdot \mid y)$ in~\citet{trippe2023diffusion}.
Instead, we implement a Markov kernel $\mathcal{K}$ keeping the filtering posterior $\bbB(u_{1:K} \mid v_{0:K}, u_0)$, for given $v_{0:K}$ and $u_0$, invariant within the Gibbs sampler, resulting in the following structure:
\begin{mdframed}[style=enumFrame]
\begin{enumerate}
    % \item sample $(Y^{j+1}_{t_k})_{k=0}^K, (Y^{j+1}_{t_k})^{j+1} \sim \bbF(\cdot \mid X^j_{0}, Y_0=y)$; \label{step:forward-discrete-filtering}
    \item sample $\bigl(Y^{j+1}_{1:K}, U^{j+1/2}_{K:0} \bigr) \sim \bbF(\cdot \mid x_0=X^j_{0}, y_0=y)$; \label{step:forward-discrete-filtering-markov}
    % \item sample $X^{j+1}_{0} = U_{T}^{j+1} \sim \bbB(\cdot \mid V_{t'_k} = Y^{j+1}_{t_k},\, k=0, \ldots, K)$. \label{step:backward-discrete-filtering}
    \item sample $X^{j+1}_{K:0} = U^{j+1}_{0:K} \sim \mathcal{K}(\cdot \mid X^{j+1/2}_{0:K})$, targeting $\bbB(u_{0:K} \mid v_{0:K} = Y^{j+1}_{K:0}, u_0=X_T^{j + 1/2})$. \label{step:backward-discrete-filtering-markov}
\end{enumerate}
\end{mdframed}
The intermediary steps, including $X^{j+1}_{1:K}$ can then be discarded.

\begin{remark}
    In practice, an additional statistical bias remains. 
    Indeed, due to $\bbB$ being a learned model (as well as $\bbF$ when treating of Schr\"odinger models), sampling from $\bbF(\cdot \mid x_0=X^j_{0}, y_0=y)$ is not equivalent to sampling from $\bbB(\cdot \mid x_0=X^j_{0}, y_0=y)$. 
    Thankfully, this can be corrected too, and we detail the procedure in Appendix~\ref{sec:app-gibbs-correction}.
\end{remark}

A natural algorithm for this purpose is given by \emph{conditional sequential Monte Carlo}~\citep[CSMC,][]{andrieu2010particle}, which modifies the standard particle filter algorithm into a Markov kernel.
CSMC is known to be fast converging under weak conditions and scales well to long time horizons~\citep{andrieu2018uniform} as well as diminishing discretisation steps in continuous systems~\citep[provided an adequate implementation is used, see, e.g.,][]{karppinen2023bridge}, including for as little as $N> 2$ Monte Carlo samples~\citep[]{lee2020coupled}.
We assume that the conditional transition density $b^V_{k}$ of $\bbB_{k}$ can be evaluated, and we provide pseudo-code for the CSMC algorithm (as well as the unconditional version, which we use later in Section~\ref{sec:separable}) applied to step~\ref{step:backward-discrete-filtering-markov} above in Algorithm~\ref{alg:csmc}.
For more details of the algorithm, we refer to~\citet[Chaps. 10 and 15]{chopin2020book} and~\citet{andrieu2010particle}.
In practice, to increase statistical efficiency, we use lower-variance resampling schemes and a final index selection step as in~\citet{karppinen2023bridge,chopin2015particlev1}.

\begin{remark}
    When we are willing to make the assumption that $\pi_{T} \equiv \pi_{\mrm{ref}}$, then an alternative is to initialise Algorithm~\ref{alg:csmc} with $U_0^n \sim \pi_{\mrm{ref}}$ rather than starting at $k=1$ from $\bbB_{0}(\cdot \mid y^j_K, x^{j}_K)$.
\end{remark}

\begin{remark}
    The computational complexity of the method is exactly that of a particle filter using the same number of particles, and we therefore do not pay any additional cost compared to \citet{trippe2023diffusion} when running the method for a \emph{single} iteration of our Markov chain. 
    On the other hand, the memory cost would generally scale as $O(T \times N)$, due to the need to have access to $x^{j}_{0:K}$. 
    In Section~\ref{subsubsec:tractable-forwards-dynamics}, we discuss how this memory cost can be fully mitigated when the forward noising SDE is separable and Gaussian.
\end{remark}

\begin{algorithm}
    \caption{(Un)conditional particle filter.}
    \label{alg:csmc}
    \begin{algorithmic}[1]
        \INPUT{The current Markov chain state $x^{j}_{0:K}$, measurement path $y^j_{0:K}$, number of particles $N > 1$, and conditional flag $F \in \{0, 1\}$. 
        Note that we shorten $n\!\upharpoonright_{1}^N$ as ``$n=1,\ldots,N$''.}
        \OUTPUT{An updated state $x^{j+1}_{0}$, and a marginal likelihood estimator $Z^j$.}
        \STATE{Sample $\lbrace U^{n}_{1} \rbrace_{n=1}^N \sim \bbB_{0}(\cdot \mid y^j_K, x^{j}_K)$, and set $Z^j=0$}
        \IF{$F = 1$}
            \STATE{Sample $n^*_0 \sim \mathcal{U}[1, N]$, and set $U^{n^*_0}_{0} = x^j_{K}$.}
        \ENDIF
        \STATE{Set $\lbrace w^n_1 = b^V_{1}(y^{j}_{K-2} \mid y^{j}_{K-1}, U^{n}_{1}) \rbrace_{n=1}^N$ and $\ell_1 = \sum_{n=1}^N w_1^n$}
        \STATE{Normalise $w^n_1 = w^n_1 / \ell_1$ for $n\!\upharpoonright_{1}^N$, and $Z^j =  \frac{Z \, \ell_1}{N}$}
        \FOR{$k = 2, \ldots, K$}
            % \COMMENT{Conditional resampling}
            \STATE{Sample $A^n_k = i$ with probability $w^i_{k-1}$ for $n\!\upharpoonright_{1}^N$\label{line:resampling}}
            \STATE{Sample $U_{k}^n \sim \bbB_{k}(\cdot \mid y^j_{K-k}, U^{A^n_{k-1}}_{k-1})$ for $n\!\upharpoonright_{1}^N$}
            \IF{$F=1$}
                \STATE{Sample $n_k^* \sim \mathcal{U}[1, N]$, and set $U^{n^*_k}_{k} = y^j_k$ and $A^{n^*_k}_k = n^*_{k-1}$}
            \ENDIF
        \STATE{Set $\lbrace w^n_k = b^v_{k}(y^{j}_{K-k-1} \mid y^{j}_{K-k}, U^{n}_{k}) \rbrace_{n=1}^N$ and $\ell_k = \sum_{n=1}^N w_k^n$}
        \STATE{Normalise $w^n_k = w^n_k / \ell_k$ for $n\!\upharpoonright_{1}^N$, and $Z^j =  \frac{Z \, \ell_k}{N}$}
        \ENDFOR
        \STATE{Sample $U_{K}^n \sim \bbB_{K-1}(\cdot \mid y^j_{K-1}, U^{A^n_{K-1}}_{K-1})$ for $n\!\upharpoonright_{1}^N$}
        \IF{$F=1$}
            \STATE{Sample $n_k^* \sim \mathcal{U}[1, N]$, and set $U^{n^*_K}_{K} = x_0^j$ and $A^{n^*_k}_k = n^*_{k-1}$}
        \ENDIF
        \STATE{Sample $B_K \in \lbrace 1,\ldots,N \rbrace$ with probabilities $\lbrace w^n_{K-1} \rbrace_{n=1}^N$ and set $x^{j+1}_{0} = U^{B_K}_{K}$}
    \end{algorithmic}
\end{algorithm}

    \section{SEPARABLE DYNAMICS}\label{sec:separable}
    The approach described in the previous section is generic, and in particular, it is applicable to non-separable noising SDEs, that is, those that have drifts $\mu^X$ and $\mu^Y$ that both depend on $X$ and $Y$, such as the SDEs resulting from Schr\"odinger bridges.
However, many diffusion models are derived from a separable forward noising SDE, typically given as an Ornstein--Uhlenbeck process, where $\mu^X$ only depends on $X$ and $\mu^Y$ on $Y$.
These models take the form
\begin{equation}\label{eq:separable}
\dd \begin{bmatrix}
    Y_t \\ X_t
\end{bmatrix} = A_t \begin{bmatrix}
    Y_t \\ X_t
\end{bmatrix} \dd t + \Sigma_t \dd W_t    
\end{equation}
with block diagonal $\Sigma_t$ and $A_t$. 
This is particularly favourable for image inpainting problems to leverage pre-trained diffusion models, where $x$ and $y$ stand for the missing and observed parts, repectively, of an image.
In this section, we describe how our approach can be further improved or adapted for these models.

\subsection{IMPROVING THE EFFICIENCY OF THE CSMC}\label{subsubsec:tractable-forwards-dynamics}
In the case when the forward-noising dynamics are given by an SDE of the form~\eqref{eq:separable}, the conditional trajectories $\bbF(\cdot \mid x_T, y_T, x_0, y_0)$ are tractable, which allows improving the procedure outlined in Section~\ref{sec:fwd-bwd-conditioning} by implementing step~\ref{step:forward-discrete-filtering} as simulating a diffusion bridge.
Namely, instead of simulating the dynamics $\bbF(\cdot \mid X_{0}^j, y)$ forward and storing the result to then feed the CSMC sampler, we can obtain $X^j_T, Y_T^j$ either from the forward simulation (or from $\pi_{\mrm{ref}}$ if it is trusted) and then implement the bridge backwards in time, simulating $\bbF(x_{k}, y_{k} \mid x^{j}_{k+1}, y^{j}_{k+1}, x_0, y_0)$ online rather than storing the full path.
This is particularly useful for high-dimensional distributions for which samples are expensive to store.

\begin{remark}
    We note that this can be done approximately in the context of non-separable nonlinear SDEs~\citep{bladt2016simulation}, for example, arising from Schr\"odinger bridges. 
    However, since we specifically focus on exact conditioning, we elect not to treat this case here.
\end{remark}

\subsection{A PSEUDO-MARGINAL IMPLEMENTATION}\label{subsec:separable-case}
When the forward noising process is given as in \eqref{eq:separable}, we can implement a pseudo-marginal counterpart of Section~\ref{sec:fwd-bwd-conditioning}, known as the particle Markov chain Monte Carlo (PMCMC) method, or more precisely the particle marginal Metropolis--Hastings (PMMH) algorithm~\citep{andrieu2010particle}.
This approach relies on the fact that particle filtering returns an unbiased estimate of the marginal likelihood of the path $Y_{1:K}$ under $\bbB$, given as $\prod_{k=1}^K \frac{1}{N}\sum_{n=1}^N w_k^n$, outputted by Algorithm~\ref{alg:csmc} (with $F=0$), which can be used in a Metropolis--Hastings step to sample from the posterior distribution of the path $X_{0:K}$.
In the context of our problem, this corresponds to the following procedure. 
Given $X^{j}_0$ distributed according to $\pi(\cdot \mid y)$ and an unbiased estimator of the marginal likelihood $Z^j$ of $\bbB(y_{K:0}) = \int \bbB(y_{K:0}, \dd u_{0:K})$:
\begin{mdframed}[style=enumFrame]
\begin{enumerate}
    \item propose a path $Y^{*}_{1:K}$ from a proposal distribution $q(\cdot \mid Y^j_{0:K})$ keeping $\bbF(\dd y_{1:K} \mid y_0=y)$ invariant
    \footnote{It is possible to use other proposals, not reversible with respect to $\bbF(\cdot \mid y_0)$, but this would require modifying the acceptance probability in step~\ref{item:sample-forward-path-accept} to account for the difference introduced by the proposal.};
    \label{item:sample-forward-path-sep}
    \item run Algorithm~\ref{alg:csmc} (with $F=0$) targeting $\bbB(\cdot \mid Y^{*}_{K:0})$ to obtain a marginal likelihood estimate $Z^*$ and a proposed $X^{*}_{0}$;
    \item set $\alpha = \frac{Z^*}{Z^j}$; \label{item:sample-forward-path-accept}
    \item set $X^{j+1}_0 = X^*_0$, $Z^{j+1} = Z^*$, and $Y^{j+1}_{1:K} = Y^{*}_{1:K}$ with probability $1 \wedge \alpha$, otherwise set $X^{j+1}_0 = X^j_0$, $Z^{j+1} = Z^j$, and $Y^{j+1}_{1:K} = Y^{j}_{1:K}$.
\end{enumerate}
\end{mdframed}
The algorithm above marginally keeps $\pi(\cdot \mid y)$ invariant under the sole assumption that $\pi_T = \pi_{\mrm{ref}}$. %\zz{I changed the algo above, double check}

\begin{remark}\label{rem:pcn}
    In step~\ref{item:sample-forward-path-sep}, we take an $\bbF(\cdot \mid y_0)$-invariant kernel.
    In this work, we choose the preconditioned Crank--Nicolson~\citep[PCN,][]{cotter2013crank} proposal, which is well adapted to Gaussian priors such as \eqref{eq:separable}.
    Noting that $\bbF(y_{1:K} \mid y_0=y)$ is a linear transformation of Gaussian variables $\eta$, and writing $\eta^j$ for those corresponding to $Y^{j}_{1:K}$,
    this consists in sampling $q$ as follows: given a step size $\delta > 0$, (i) sample $\eta' \sim \mathcal{N}(0, I)$, (ii) set $\eta^* = \frac{2}{2 + \delta} \, \eta^j + \sqrt{1 - \frac{4}{4 + 4 \delta + \delta^2}} \, \eta'$, and (iii) form $Y^*$ from $\eta^*$.
    Importantly, this can be done with zero memory cost if one has access to the random number generator that was used to sample $\eta^j$ originally, in which case we can simply re-simulate $\eta^j$ alongside $\eta'$ at zero memory cost and negligible (compared to evaluating the SDE drift) computational cost.
\end{remark}

\begin{remark}\label{rem:pcn-tunning}
    The parameter $\delta$ represents an exploration-exploitation trade-off: $\delta$ small corresponds to slow but likely successful exploration of the state-space, $\delta$ large corresponds to fast but likely unsuccessful exploration.
    In practice, for $\delta=0$ (which would result in a non-ergodic algorithm), the acceptance rate is fully dictated by the pseudo-marginal, in part by the variance of the normalising constant estimates, and in part by pseudo-marginal MCMC methods not recovering perfect samplers~\citep[Appendix F]{andrieu2018uniform}
    For a large $\delta$, the acceptance will be fully dominated by the fact that the two normalising constants for two different paths $Y^k_{0:T}$ and $Y^*_{0:T}$ will be very different making the chain get stuck.
    The $\delta$ parameter interpolates between these two regimes, whereby we need to correlate $y_{0:T}$ and $y'_{0:T}$ so that we can get close enough to the ``ideal'' scenario of the sampler being almost fully dominated by the pseudo-marginal behaviour.
\end{remark}

    \section{Experiments}\label{sec:experiments}
    In this section, we conduct quantitative and qualitative experiments to validate our proposed method (i.e., Gibbs-CSMC in Sections~\ref{subsec:gibbs-sampler} and~\ref{subsubsec:tractable-forwards-dynamics} and PMCMC in Section~\ref{subsec:separable-case}) on synthetic and real data. 
We compare to related methods: the standard particle filter~\citep[PF,][]{trippe2023diffusion, dou2024diffusion}, twisted particle filter~\citep[TPF,][]{wu2023practical}, and conditional score matching~\citep[CSGM,][]{song2021scorebased}. 
In all experiments, particle filters use stratified resampling, and the conditional particle filter uses the conditional ``killing'' resampling of~\citet{karppinen2023bridge} instead of the multinomial resampling in Algorithm~\ref{alg:csmc}.
Unless otherwise mentioned, we repeat every experiment 100 times independently, reporting the mean, standard deviation, and/or quantile of their results. Our implementations are available at \url{\codeaddress}.

\subsection{SYNTHETIC CONDITIONAL SAMPLING}
\label{sec:experiment-gp-regression}
In our first ablation experiment, we evaluate the convergence of the conditioning methods to a tractable conditional distribution.
Here, the target $\pi(x \mid y)$ is the posterior of a Gaussian process regression model with known ground truth.
We use a linear forward noising process, so that the reverse SDE~\eqref{eq:bwd-sde-base} is computed exactly, isolating the error coming from the conditioning methods only. 
For each method, we generate 10,000 posterior samples, and assess the errors using Kullback--Leibler (KL) divergence, Wasserstein--Bures (Bures) distance, and mean absolute errors of the marginal means and variances. 
Details are given in Appendix~\ref{appendix:gp}.

\begin{table*}[t!]
    \caption{Errors statistics for the the conditional sampling problem of Section~\ref{sec:experiment-gp-regression}. The left and right panels show the errors when using 10 and 100 particles, respectively. The columns ``Mean'' and ``Variance'' show the errors in the marginal mean and variance of the posterior. The number after PMCMC is the PCN parameter $\delta$ applied. Since CSGM is computed exactly, it here serves as the reference and is not for comparisons.}
    \label{tbl:toy-errs}
    \centering
    \resizebox{.8\linewidth}{!}{%
        \begin{tabular}{@{}lllll:llll@{}}
            \toprule
            & KL               & Bures           & Mean ($\times 10^{-2}$) & Variance ($\times 10^{-2}$) & KL               & Bures           & Mean ($\times 10^{-2}$) & Variance ($\times 10^{-2}$) \\ \midrule
            PF          & $1.67 \pm 0.12$  & $0.71 \pm 0.10$ & $4.51 \pm 0.74$         & $4.02 \pm 0.05$              & $1.12 \pm 0.07$  & $0.29 \pm 0.05$ & $3.12 \pm 0.49$         & $2.33 \pm 0.05$              \\
            Gibbs-CSMC  & $\mathbf{1.05} \pm 0.01$  & $\mathbf{0.07} \pm 0.00$ & $\mathbf{0.88} \pm 0.08$         & $\mathbf{0.41} \pm 0.03$              & $\mathbf{0.76} \pm 0.00$  & $\mathbf{0.04} \pm 0.00$ & $\mathbf{0.54} \pm 0.04$         & $\mathbf{0.32} \pm 0.02$              \\
            PMCMC-0.005 & $26.31 \pm 1.41$ & $1.22 \pm 0.09$ & $4.45 \pm 0.45$         & $1.74 \pm 0.12$              & $22.11 \pm 1.04$ & $1.18 \pm 0.08$ & $5.01 \pm 0.47$         & $1.73 \pm 0.11$              \\
            PMCMC-0.001 & $6.83 \pm 0.57$  & $0.50 \pm 0.07$ & $3.48 \pm 0.39$         & $1.24 \pm 0.08$              & $4.50 \pm 0.21$  & $0.55 \pm 0.07$ & $4.68 \pm 0.45$         & $1.50 \pm 0.07$              \\
            TPF         & $5.22 \pm 0.67$  & $1.32 \pm 0.56$ & $6.04 \pm 1.92$         & $2.26 \pm 0.47$              & $2.61 \pm 0.34$  & $0.78 \pm 0.32$ & $5.84 \pm 0.02$         & $1.09 \pm 0.50$              \\\cdashlinelr{1-9}
            CSGM-exact  & $0.66 \pm 0.01$  & $0.03 \pm 0.00$ & $0.33 \pm 0.04$         & $0.20 \pm 0.02$              & $\leftarrow$             & $\leftarrow$            & $\leftarrow$                    & $\leftarrow$                         \\ \bottomrule
        \end{tabular}
    }
\end{table*}

As shown by Table~\ref{tbl:toy-errs}, Gibbs-CSMC outperforms all other methods for all metrics, effectively correcting the biase of the PF approach~\citep{trippe2023diffusion}.
Our PMCMC approach also improves on PF but is sensitive to the calibration of the PCN parameter $\delta$: errors decrease when reducing $\delta=0.005$ to $\delta=0.001$.
While PMCMC shows good marginal mean and variance errors, its KL performance suggests poor approximation to the off-diagonal GP posterior covariance. 
TPF performs worse than PF with higher standard deviations. Increasing the number of particles improves errors across all methods.
More qualitative results are given in Appendix~\ref{appendix:gp}.

\begin{figure}[t!]
    \centering
    \includegraphics[width=.9\linewidth]{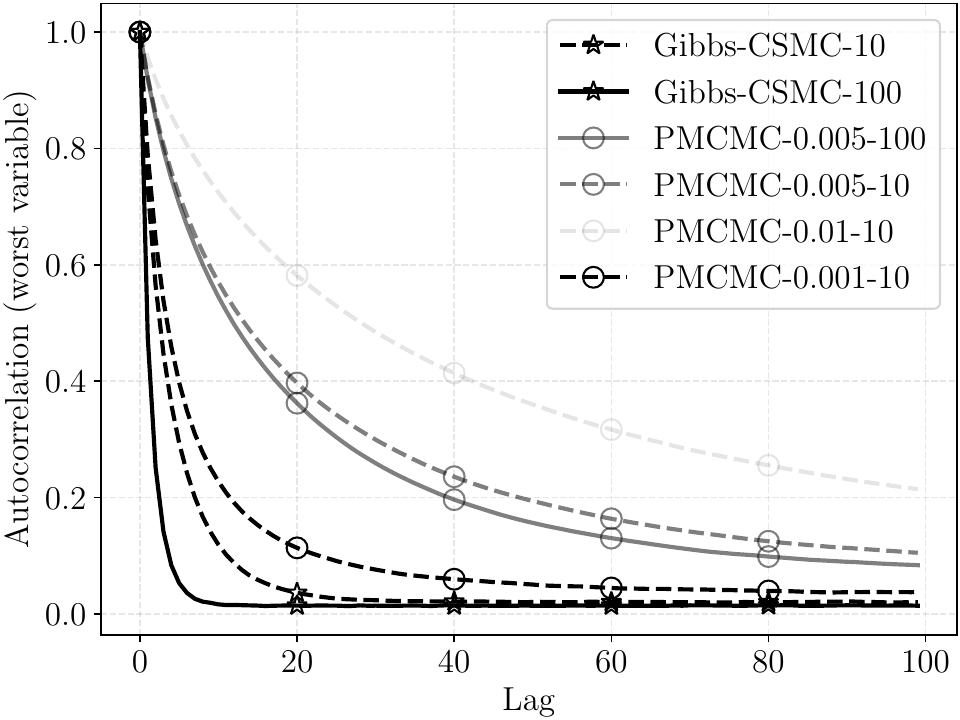}
    \vspace{-15pt}
    \caption{Autocorrelations (of the worst dimension) in Section~\ref{sec:experiment-gp-regression}. 
    For each of the 100 dimensions the autocorrelation is averaged over four chains. 
    We see that Gibbs-CSMC outperforms PMCMC, with higher particle counts reducing autocorrelation.}
    \label{fig:toy-autocorr}
\end{figure}

Figure~\ref{fig:toy-autocorr} shows the autocorrelations of our Gibbs-CSMC and PMCMC methods. 
Gibbs-CSMC exhibits better autocorrelation than PMCMC, even with fewer particles.
PMCMC's autocorrelation improves by decreasing the PCN parameter, mostly due to the variance of the log-likelihood estimation (see Section~\ref{subsec:separable-case}) of the PMCMC method~\citep{andrieu2009pseudomarginal,doucet2015efficient}, sometimes preventing the algorithm from moving, and which is negatively affected by $\delta$. 
Lastly, the figure also confirms that increasing the number of particles improves the autocorrelations overall.

\begin{figure}[t!]
    \centering
    \includegraphics[width=.9\linewidth]{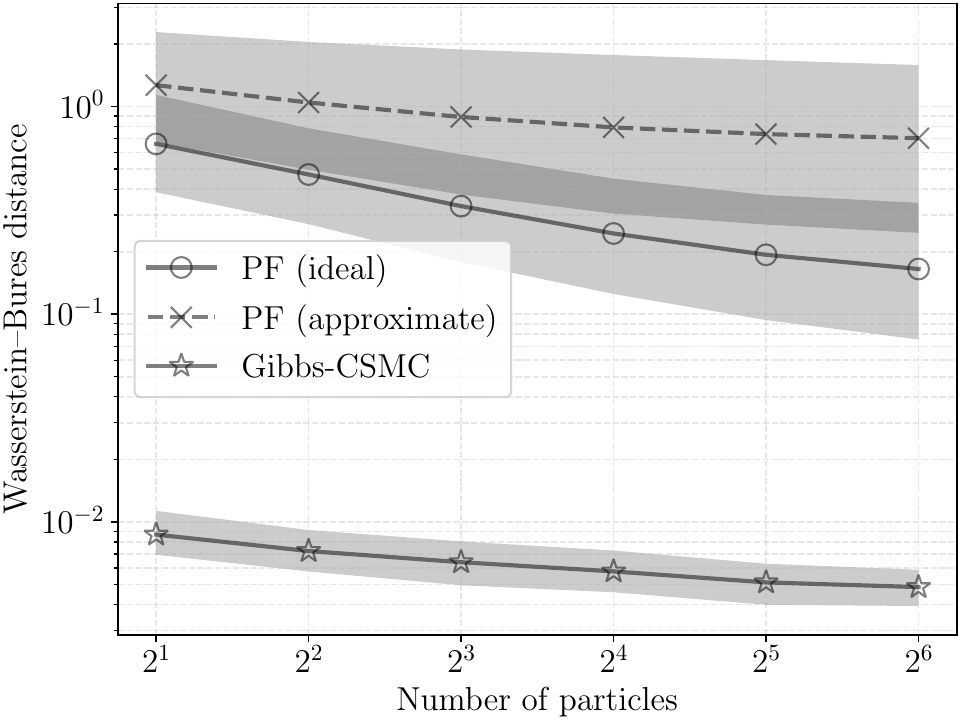}
    \vspace{-15pt}
    \caption{Conditional sampling errors on a Gaussian Schr\"{o}dinger bridge: PF (ideal) uses the exact posterior (generally unavailable), while PF (approximate) uses a standard Normal. Gibbs sampling significantly improves quality, even over PF (ideal), underscoring PF's inaccuracies.}
    \label{fig:sb-gaussian}
\end{figure}

\subsection{CONDITIONAL SAMPLING ON NON-SEPARABLE NOISING PROCESSES}
\label{sec:experiment-sb}
A key merit of our method is generalisability: it works with both separable and non-separable noising processes. 
Precisely, it does not need $X_t$ and $Y_t$ of the noising process in \eqref{eq:forward-diffusion} to be independent given $X_0$ and $Y_0$.
In contrast, standard PF approach of~\citet{dou2024diffusion, trippe2023diffusion} relies on separability.
For non-separable processes, their method requires sampling $X_0$ to generate a path $t\mapsto Y_t$, ideally from the true conditional distribution, ideally from the true $X_0 \sim \pi(x \mid y)$, but in practice, approximate samplers are used, introducing biases that are difficult to remove.

To quantify the bias from the non-separability, we perform conditional sampling on a Gaussian Schr\"{o}dinger bridge model~\citep[for which closed-form expressions are available, see][]{Bunne2023} using our Gibbs-CSMC and the PF of~\citet{trippe2023diffusion}. 
We do not include PMCMC given its non-applicability here~(see Section~\ref{subsec:separable-case}).
Continuing with the GP regression problem from Section~\ref{sec:experiment-gp-regression}, we replace the reference $\mathcal{N}(0, I)$ with $\mathcal{N}(0, \Sigma)$, where $\Sigma$ is a random covariance matrix from a Wishart distribution, to highlight the impact of non-separability.
Results in Figure~\ref{fig:sb-gaussian} show that approximating the initial $\pi(x \mid y)$ leads to significant errors compared to the PF using exact initial sampling. 
Additionally, increasing the number of particles does not mitigate this initial bias, underscoring the shortcomings of the standard approach in non-separable models.

\begin{figure}[t!]
    \centering
    \includegraphics[width=\linewidth]{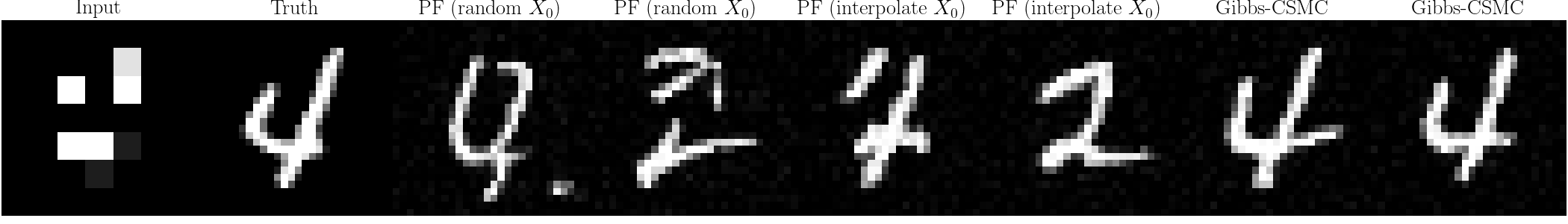}
    \caption{MNIST super-resolution (x4) on a non-separable Schr\"{o}dinger bridge noising process. Each method shows two samples (more are in Figure~\ref{fig:app-sb-imgs}). 
    We find that PF is significantly affected by its initialisation of $X_0$, while Gibbs-CSMC is largely unaffected.}
    \label{fig:sb-imgs-examples}
\end{figure}

We also qualitatively assess the MNIST super-resolution on a pre-trained Schr\"{o}dinger bridge. 
This time, we lack an exact $X_0$ sampler for the PF method, and we thus use a uniform random $X_0$, and an $X_0$ using linear interpolation of $y_0$. 
Results in Figure~\ref{fig:sb-imgs-examples} show that the initial $X_0$ in the forward noising process significantly affects the outcomes. 
Using uniform $\mathcal{U}(0, 1)$ initialisation leads to substantial deviations from the truth. 
In contrast, linear interpolation, a better initialisation, yields results that are closer to the truth, though the quality remains poor. 
Conversely, Gibbs-CSMC does not have the initialisation issue and recovers the original digit as the MCMC chain converges. 
More results are given in Appendix~\ref{sec:app-sb}.

\subsection{INPAINTING AND SUPER-RESOLUTION}
\label{sec:experiment-imgs}
In this section, we apply our methods for inpainting and super-resolution on MNIST (resolution 28 by 28) and CelebA-HQ (resolution 64 by 64) images. 
For a complete description of the inpainting and super-resolution problems and additional experimental details, we refer the readers to Appendix~\ref{app:imgs-samples}.
Given pre-trained~\citep[via the standard score matching method of][]{song2021scorebased} unconditional generative diffusions for MNIST and Celeba-HQ images, we then perform conditional sampling of missing (inpainting) or deblurred (super-resolution) of images given an observation thereof.
To quantify the results, for each test image we recover 100 conditional image samples, and we compute the peak signal-to-noise ratio (PSNR), structural similarity index (SSIM), and learnt perceptual patch similarity~\citep[LPIPS,][]{zhang2018cvpr} metrics averaged over the 100 samples.

\begin{table*}[t!]
    \caption{Results of MNIST inpainting and super-resolution. In the left and right panels, we show the results using 10 and 100 particles, respectively. Bold numbers are the best column-wise.}
    \label{tbl:mnist}
    \centering
    \resizebox{.85\linewidth}{!}{%
        \begin{tabular}{@{}lllll:llll@{}}
            \toprule
            & \multicolumn{4}{c}{10 particles} & \multicolumn{4}{c}{100 particles} \\
            \cmidrule(l){2-5} \cmidrule(l){6-9}
            \multirow{2}{*}{MNIST} & \multicolumn{2}{l}{Inpainting} & \multicolumn{2}{l}{Super-resolution} & \multicolumn{2}{l}{Inpainting} & \multicolumn{2}{l}{Super-resolution} \\ \cmidrule(l){2-5} \cmidrule(l){6-9}
            & PSNR  & SSIM & PSNR  & SSIM & PSNR  & SSIM & PSNR  & SSIM \\ \midrule
            PF         & $15.85 \pm 4.5$ & $0.74 \pm 0.2$ & $10.14 \pm 1.8$ & $0.32 \pm 0.2$ & $17.34 \pm 4.5$ & $0.79 \pm 0.1$ & $11.66 \pm 2.4$ & $0.45 \pm 0.2$ \\
            Gibbs-CSMC & $16.79 \pm 4.6$ & $0.77 \pm 0.1$ & $10.84 \pm 1.8$ & $0.38 \pm 0.2$ & $17.82 \pm 4.1$ & $\mathbf{0.82} \pm 0.1$ & $14.06 \pm 3.4$ & $0.61 \pm 0.2$ \\
            PMCMC-0.005      & $\mathbf{17.51} \pm 4.4$ & $\mathbf{0.80} \pm 0.1$ & $\mathbf{12.88} \pm 2.9$ & $\mathbf{0.54} \pm 0.2$ & $\mathbf{17.84} \pm 4.5$ & $0.81 \pm 0.1$ & $\mathbf{14.22} \pm 2.8$ & $\mathbf{0.63} \pm 0.2$ \\
            TPF        & $13.45 \pm 3.1$ & $0.48 \pm 0.2$ & $9.67 \pm 1.6$  & $0.19 \pm 0.1$ & $13.89 \pm 3.2$ & $0.51 \pm 0.2$ & $10.38 \pm 2.0$ & $0.25 \pm 0.1$ \\
            CSGM       & $15.17 \pm 4.2$ & $0.71 \pm 0.2$ & $9.72 \pm 1.6$  & $0.28 \pm 0.1$ & $\leftarrow$  & $\leftarrow$ & $\leftarrow$  & $\leftarrow$ \\ \bottomrule
        \end{tabular}
    }

    \vspace*{5pt}
    \caption{Results of CelebA-HQ inpainting and super-resolution. In the left and right panels, we apply 2 and 10 particles, respectively. Bold numbers are the best column-wise.}
    \label{tbl:celeba}
    \centering
    \resizebox{.99\linewidth}{!}{%
        \begin{tabular}{@{}lllllll:llllll@{}}
            \toprule
            & \multicolumn{6}{c}{2 particles} & \multicolumn{6}{c}{10 particles} \\
            \cmidrule(l){2-7} \cmidrule(l){8-13}
            \multirow{2}{*}{CelebA-HQ} & \multicolumn{3}{l}{Inpainting} & \multicolumn{3}{l}{Super-resolution} & \multicolumn{3}{l}{Inpainting} & \multicolumn{3}{l}{Super-resolution} \\ \cmidrule(l){2-7} \cmidrule(l){8-13}
            & PSNR  & SSIM & LPIPS & PSNR  & SSIM & LPIPS & PSNR & SSIM & LPIPS & PSNR & SSIM & LPIPS \\ \midrule
            PF         & $22.88 \pm 3.7$ & $0.85 \pm 0.05$ & $0.05 \pm 0.03$ & $23.29\pm1.8$ & $0.79 \pm 0.05$ & $0.11\pm0.05$ & $23.98\pm3.6$ & $0.870.05$ & $\mathbf{0.04}\pm0.02$ & $24.57\pm1.7$ & $0.82\pm0.04$ & $0.09\pm0.04$     \\
            Gibbs-CSMC & $22.86 \pm 3.7$ & $0.86 \pm 0.05$ & $0.05 \pm 0.03$ & $23.75\pm1.8$ & $0.80\pm0.05$ & $0.11\pm0.06$  & $24.22\pm3.5$ & $0.87\pm0.04$ & $\mathbf{0.04}\pm0.03$ & $25.07\pm1.8$ & $0.84\pm0.04$ & $0.09\pm0.05$      \\
            PMCMC-0.005      & $\mathbf{23.96} \pm 3.5$ & $\mathbf{0.87} \pm 0.05$ & $\mathbf{0.04}\pm0.03$ & $\mathbf{24.37}\pm1.7$ & $\mathbf{0.82}\pm0.04$ & $\mathbf{0.10}\pm0.04$  & $\mathbf{24.68}\pm 3.5$ & $\mathbf{0.88}\pm0.04$ & $\mathbf{0.04}\pm0.02$ & $\mathbf{25.30}\pm1.8$ & $\mathbf{0.85}\pm0.04$ & $\mathbf{0.07}\pm0.03$      \\
            TPF        & $13.79\pm2.4$ & $0.59\pm0.06$ & $0.20\pm0.07$ & $12.71\pm1.6$ & $0.44\pm0.08$ & $0.23\pm0.10$  & $13.93\pm2.4$ & $0.58\pm0.06$ & $0.20\pm0.07$ & $12.92\pm1.6$ & $0.45\pm0.08$ & $0.23\pm0.10$      \\
            CSGM       & $22.86\pm3.6$ & $0.85\pm0.05$ & $0.05\pm0.03$ & $23.48\pm1.8$ & $0.79\pm0.05$ & $0.11\pm0.05$  & $\leftarrow$  & $\leftarrow$  &  $\leftarrow$ &  $\leftarrow$  &  $\leftarrow$ &  $\leftarrow$  \\ \bottomrule
        \end{tabular}
    }
\end{table*}

The results are summarised in Tables~\ref{tbl:mnist} and~\ref{tbl:celeba}. 
We see that our PMCMC approach is the best in terms of all the metrics, followed by Gibbs-CSMC. 
Importantly, our PMCMC and Gibbs-CSMC methods outperform the PF approach, showing that our MCMC framework indeed compensates the bias exhibited by PF for real applications. 
The classical CSGM approach performs similarly as PF, however, the filtering-based approaches can be systematically improved by increasing the number of particles, while CSGM cannot~

\begin{figure*}
    \centering
    \includegraphics[width=.245\linewidth]{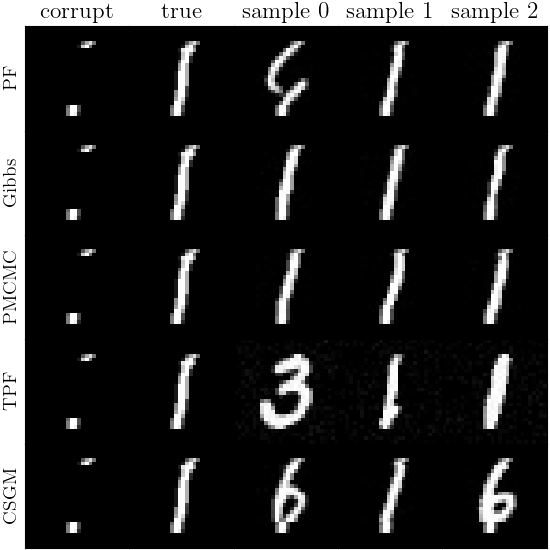}
    \includegraphics[width=.245\linewidth]{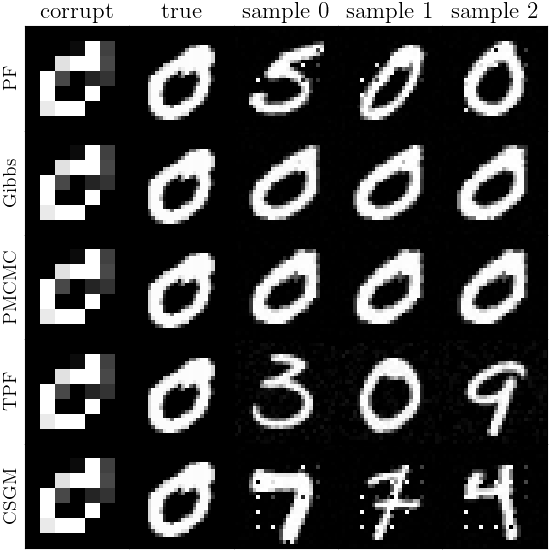}
    \includegraphics[width=.245\linewidth]{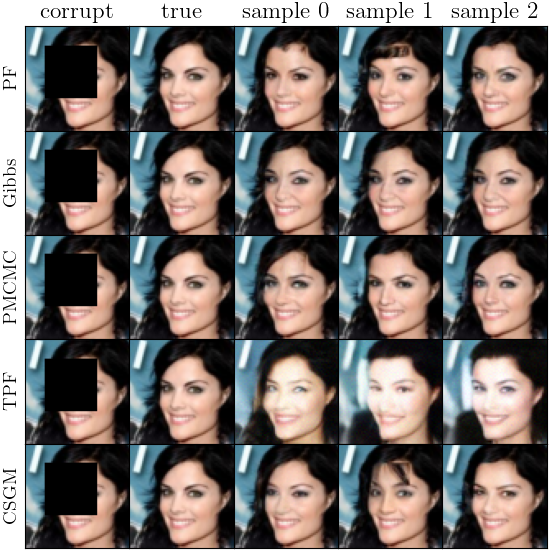}
    \includegraphics[width=.245\linewidth]{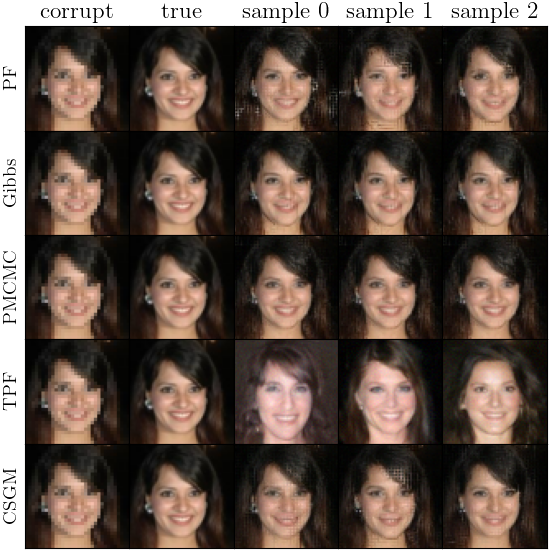}
    \vspace{-15pt}
    \caption{Examples of inpainting (first and third panels) and super-resolution (second and fourth panels) on MNIST and CelebA-HQ. In each panel, the first to the last rows show the results of PF, Gibbs-CSMC, PMCMC-0.005, TPF, and CSGM, respectively. We see that the samples generated by Gibbs-CSMC and PMCMC have overall better quality.}
    \label{fig:imgs-examples}
\end{figure*}

In Figure~\ref{fig:imgs-examples}, we illustrate the methods for inpainting and super-resolution on a single task.
We see that PF, TPF, and CSGM can generate unrealistic samples, and the super-resolution samples are sometimes contaminated by pixel distortions (e.g., the isolated white dots in the second panel, first row, third-fifth columns in the figure). 
This is particularly true for TPF despite it largely higher computational cost~\citep[due to taking gradient with respect to the score function in its twisting function,][]{wu2023practical} as it requires to further approximate the likelihood model $\pi(y \mid x)$ by a smoother version to be usable.
However, our Gibbs-CSMC and PMCMC methods are able to produce better quality samples, albeit more correlated ones, as those wrong samples are often rejected by the MCMC kernel. 
For more examples and results, we refer the readers to Appendix~\ref{app:imgs-samples}.

    \section{RELATED WORKS}\label{sec:related}
    In the past few years, several methods have been proposed to perform conditional sampling within diffusion models~\citep[see e.g., a review in][]{Zhao2024rsta}.
    The first way to do so was proposed alongside score-based denoising diffusion models in the landmark work by~\citet{song2021scorebased}.
    They rely on learning an approximate conditioning drift model \emph{after} the fact, adding a layer of approximate learning on top of the unconditional one~\citep[see typical approximations in][]{chung2023diffusion, song2023pseudoinverseguided, mardani2024a, denker2024deft}.
    This type of idea can also be found in~\citet{Shi2022conditional} who train a Schr\"odinger bridge on the joint distribution of the latent variable $X$ and the observed $Y$ when the model $\pi(y \mid x)$ is available.
    Closer to our work, \citet{wu2023practical} propose using a twisted particle filter~\citep{Whiteley2014twistedPF} to efficiently bridge between the reference distribution and the conditional $\pi(x \mid y)$ at the cost of additional expensive computations in the procedure.
    Finally, \citet{trippe2023diffusion}, as well as the contemporaneous~\citet{dou2024diffusion}, express the problem of conditioning as a bridge for simulated observations too.
    In fact, \citet{trippe2023diffusion} is exactly given as a special case of Section~\ref{subsec:separable-case} where the new path is proposed independently from the previous one (i.e., in our formulation, $q \equiv \bbF(\cdot \mid y_0)$), and the resulting sample is accepted unconditionally (i.e., their method is equivalent to setting $\alpha = 1$), ignoring the fact that two $Y$ paths sampled under the forward dynamics (the prior) may have different marginal likelihoods under the backward denoising model.
    This means that their method is inherently biased and does not asymptotically target the right distribution, despite their consistency result, even under the assumption $\pi_T \equiv \pi_{\mrm{ref}}$. 
    The resulting bias is reflected in our experiments.
    This defect is shared by~\citet{dou2024diffusion}. 
    Additional background details can be found in Appendix~\ref{app:background}.

    \section{CONCLUSION}\label{sec:conclusions}
    We have presented a novel MCMC-based method for sampling from conditional distributions expressed as diffusion models and Schr\"odinger bridges.
    Our method relies on formulating the sampling procedure as an inference problem on missing observation data, thanks to which we can apply classical particle MCMC algorithms~\citep{andrieu2010particle} to perform (asymptotically) exact sampling within the model.
    We have demonstrated the performance and correctness of our method on simulated and real datasets, for inpainting and super-resolution problems alike.

    One iteration of CSMC (or the pseudo-marginal version of the algorithm) has exactly the same computational cost as its unconditional version for the same number of particles. This means that the additional cost is fully encapsulated by the number of iterations required (see, e.g., Figure~\ref{fig:app-autocorr-imgs}) to achieve convergence.
    Consequently, to obtain independent samples with adequate posterior coverage, a pragmatic compromise is to repeatedly use~\citet{trippe2023diffusion} as initialisation and then correct the posterior sample by running our methods for a few iterations, preferably the CSMC version due to its improved stability. 

    While our method improves on the pre-existing methodology, in particular~\citet{trippe2023diffusion} on which we generalise, several questions remain open. 
    In particular, (i) we observed that the distribution of the particles in the backward filtering pass contracts (see, Appendix~\ref{sec:app-coalescence}) at the terminal time~\citep[this is true for all methods considered here, including]{trippe2023diffusion,wu2023practical}, and may consequently increase the autocorrelation of our MCMC chains, explaining their lack of diversity despite the overall improvement in quality.
    Because this appears throughout, it seems to be a feature of the model decomposition rather than a feature of Gibbs-CSMC or PMCMC.
    (ii) We also notice that the variance of the log-likelihood estimation in PMCMC can be large, and hence, its acceptance rate can be unstable, also making the calibration of its step size $\delta$ difficult.
    While this is detrimental to our approach, it also explains the empirical poor posterior coverage of \citet{trippe2023diffusion} who completely ignore this step.   
    Because of this, we recommend using the Gibbs-CSMC approach above PMCMC, despite the latter's occasional better experimental performance.
    
    Nevertheless, solving both these two issues of the framework is of great interest, and is likely feasible within the framework we outlined in this article by a more careful consideration of the structure of the augmented model.
    We leave these questions for future work.

    Finally, we remark that our method can straightforwardly extend to a wider class of inverse problems~\citep[see, e.g., surveys in][]{daras2024survey, huang2024diffusion, HuangYi2024}. 
    Notably, our method can, without dedicated training, solve all inverse problems with linear Gaussian likelihood models which can be represented as inpainting problems~\citep{Kawar2021}.
    Extending the method to nonlinear likelihood models, which our work does not support, would also make the method of more use to practitioners, and, as a consequence, it is a very important avenue of future work. 

\subsubsection*{Acknowledgements}
The original idea and methodology are due to AC and subsequently refined jointly by AC and ZZ. Implementation and experimental evaluation are mostly due to ZZ with help and inputs from AC. Writing was primarily done by AC with substantial help from ZZ and inputs from TS. All authors edited and validated the final manuscript. 
This work was partially supported by the Kjell och M\"{a}rta Beijer Foundation, Wallenberg AI, Autonomous Systems and Software Program (WASP) funded by the Knut and Alice Wallenberg Foundation, the project \emph{Deep probabilistic regression -- new models and learning algorithms} (contract number: 2021-04301), funded by the Swedish Research Council, and by the Research Council of Finland. The computations were enabled by the Berzelius resource provided by the Knut and Alice Wallenberg Foundation at the National Supercomputer Centre. 
Adrien Corenflos and Simo S\"arkk\"a were partially supported by the Research Council of Finland. 
Adrien Corenflos also acknowledges the financial support provided by UKRI for OCEAN (a 2023-2029 ERC Synergy grant co-sponsored by UKRI).

\bibliography{main.bib}

\begin{thebibliography}{}

\bibitem[Andrieu et~al., 2010]{andrieu2010particle}
Andrieu, C., Doucet, A., and Holenstein, R. (2010).
\newblock Particle {M}arkov chain {M}onte {C}arlo methods.
\newblock {\em Journal of the Royal Statistical Society: Series B (Statistical
  Methodology)}, 72(3):269--342.
\newblock With discussion.

\bibitem[Andrieu et~al., 2018]{andrieu2018uniform}
Andrieu, C., Lee, A., and Vihola, M. (2018).
\newblock Uniform ergodicity of the iterated conditional {SMC} and geometric
  ergodicity of particle {G}ibbs samplers.
\newblock {\em Bernoulli}, 24(2):842--872.

\bibitem[Andrieu and Roberts, 2009]{andrieu2009pseudomarginal}
Andrieu, C. and Roberts, G.~O. (2009).
\newblock The pseudo-marginal approach for efficient {Monte Carlo}
  computations.
\newblock {\em The Annals of Statistics}, 37(2):697--725.

\bibitem[Bladt et~al., 2016]{bladt2016simulation}
Bladt, M., Finch, S., and S{\o}rensen, M. (2016).
\newblock Simulation of multivariate diffusion bridges.
\newblock {\em Journal of the Royal Statistical Society Series B: Statistical
  Methodology}, 78(2):343--369.

\bibitem[Bradbury et~al., 2018]{jax2018github}
Bradbury, J., Frostig, R., Hawkins, P., Johnson, M.~J., Leary, C., Maclaurin,
  D., Necula, G., Paszke, A., Vander{P}las, J., Wanderman-{M}ilne, S., and
  Zhang, Q. (2018).
\newblock {JAX}: composable transformations of {P}ython+{N}um{P}y programs.

\bibitem[Bunne et~al., 2023]{Bunne2023}
Bunne, C., Hsieh, Y.-P., Cuturi, M., and Krause, A. (2023).
\newblock The {S}chr\"{o}dinger bridge between {G}aussian measures has a closed
  form.
\newblock In {\em Proceedings of The 26th International Conference on
  Artificial Intelligence and Statistics}, volume 206, pages 5802--5833. PMLR.

\bibitem[Cardoso et~al., 2024]{Cardoso2024monte}
Cardoso, G., el~idrissi, Y.~J., Corff, S.~L., and Moulines, E. (2024).
\newblock Monte {C}arlo guided denoising diffusion models for {B}ayesian linear
  inverse problems.
\newblock In {\em The 12th International Conference on Learning
  Representations}.

\bibitem[Chen et~al., 2021]{Chen2021optimaltransport}
Chen, Y., Georgiou, T.~T., and Pavon, M. (2021).
\newblock Optimal transport in systems and control.
\newblock {\em Annual Review of Control, Robotics, and Autonomous Systems},
  4:89--113.

\bibitem[Chopin and Papaspiliopoulos, 2020]{chopin2020book}
Chopin, N. and Papaspiliopoulos, O. (2020).
\newblock {\em An Introduction to Sequential {M}onte {C}arlo}.
\newblock Springer.

\bibitem[Chopin and Singh, 2015]{chopin2015particlev1}
Chopin, N. and Singh, S.~S. (2015).
\newblock On particle {G}ibbs sampling.
\newblock {\em arXiv preprint arXiv:1304.1887 version 1}.

\bibitem[Chung et~al., 2023]{chung2023diffusion}
Chung, H., Kim, J., Mccann, M.~T., Klasky, M.~L., and Ye, J.~C. (2023).
\newblock Diffusion posterior sampling for general noisy inverse problems.
\newblock In {\em The Eleventh International Conference on Learning
  Representations}.

\bibitem[Cotter et~al., 2013]{cotter2013crank}
Cotter, S.~L., Roberts, G.~O., Stuart, A.~M., and White, D. (2013).
\newblock {MCMC} methods for functions: modifying old algorithms to make them
  faster.
\newblock {\em Statistical Science}, 28(3):424--446.

\bibitem[Daras et~al., 2024]{daras2024survey}
Daras, G., Chung, H., Lai, C.-H., Mitsufuji, Y., Ye, J.~C., Milanfar, P.,
  Dimakis, A.~G., and Delbracio, M. (2024).
\newblock A survey on diffusion models for inverse problems.
\newblock {\em arXiv preprint arXiv:2410.00083}.

\bibitem[De~Bortoli et~al., 2021]{deBortoli2021diffusion}
De~Bortoli, V., Thornton, J., Heng, J., and Doucet, A. (2021).
\newblock Diffusion {S}chr{\"o}dinger bridge with applications to score-based
  generative modeling.
\newblock In {\em Advances in Neural Information Processing Systems},
  volume~34, pages 17695--17709.

\bibitem[Denker et~al., 2024]{denker2024deft}
Denker, A., Vargas, F., Padhy, S., Didi, K., Mathis, S., Dutordoir, V.,
  Barbano, R., Mathieu, E., Komorowska, U.~J., and Lio, P. (2024).
\newblock {DEFT}: Efficient finetuning of conditional diffusion models by
  learning the generalised $h$-transform.
\newblock {\em arXiv preprint arXiv:2406.01781}.

\bibitem[Dou and Song, 2024]{dou2024diffusion}
Dou, Z. and Song, Y. (2024).
\newblock Diffusion posterior sampling for linear inverse problem solving: a
  filtering perspective.
\newblock In {\em Proceedings of the Twelfth International Conference on
  Learning Representations}.

\bibitem[Doucet et~al., 2015]{doucet2015efficient}
Doucet, A., Pitt, M.~K., Deligiannidis, G., and Kohn, R. (2015).
\newblock Efficient implementation of {Markov chain Monte Carlo} when using an
  unbiased likelihood estimator.
\newblock {\em Biometrika}, 102(2):295--313.

\bibitem[Geman and Geman, 1984]{Geman1984stochasticrelaxation}
Geman, S. and Geman, D. (1984).
\newblock Stochastic relaxation, {G}ibbs distributions, and the {B}ayesian
  restoration of images.
\newblock {\em IEEE Transactions on Pattern Analysis and Machine Intelligence},
  6(6):721--741.

\bibitem[Heek et~al., 2023]{flax2020github}
Heek, J., Levskaya, A., Oliver, A., Ritter, M., Rondepierre, B., Steiner, A.,
  and van {Z}ee, M. (2023).
\newblock {F}lax: A neural network library and ecosystem for {JAX}.

\bibitem[Ho et~al., 2020]{Ho2020}
Ho, J., Jain, A., and Abbeel, P. (2020).
\newblock Denoising diffusion probabilistic models.
\newblock In {\em Advances in Neural Information Processing Systems},
  volume~33, pages 6840--6851.

\bibitem[Huang et~al., 2024a]{HuangYi2024}
Huang, Y., Huang, J., Liu, J., Yan, M., Dong, Y., Lv, J., Chen, C., and Chen,
  S. (2024a).
\newblock {WaveDM}: Wavelet-based diffusion models for image restoration.
\newblock {\em IEEE Transactions on Multimedia}, 26:7058--7073.

\bibitem[Huang et~al., 2024b]{huang2024diffusion}
Huang, Y., Huang, J., Liu, Y., Yan, M., Lv, J., Liu, J., Xiong, W., Zhang, H.,
  Chen, S., and Cao, L. (2024b).
\newblock Diffusion model-based image editing: A survey.
\newblock {\em arXiv preprint arXiv:2402.17525}.

\bibitem[Hyv{\"a}rinen, 2005]{hyvarinen2005score}
Hyv{\"a}rinen, A. (2005).
\newblock Estimation of non-normalized statistical models by score matching.
\newblock {\em Journal of Machine Learning Research}, 6(24):695--709.

\bibitem[Karatzas and Shreve, 1991]{Karatzas1991}
Karatzas, I. and Shreve, S.~E. (1991).
\newblock {\em Brownian Motion and Stochastic Calculus}, volume 113 of {\em
  Graduate Texts in Mathematics}.
\newblock Springer-Verlag New York, 2nd edition.

\bibitem[Karppinen et~al., 2023]{karppinen2023bridge}
Karppinen, S., Singh, S.~S., and Vihola, M. (2023).
\newblock Conditional particle filters with bridge backward sampling.
\newblock {\em Journal of Computational and Graphical Statistics}, 0(0):1--15.

\bibitem[Kawar et~al., 2021]{Kawar2021}
Kawar, B., Vaksman, G., and Elad, M. (2021).
\newblock {SNIPS}: Solving noisy inverse problems stochastically.
\newblock In {\em Advances in Neural Information Processing Systems},
  volume~34, pages 21757--21769. Curran Associates, Inc.

\bibitem[Ke et~al., 2021]{Ke2021ICCV}
Ke, J., Wang, Q., Wang, Y., Milanfar, P., and Yang, F. (2021).
\newblock {MUSIQ}: Multi-scale image quality transformer.
\newblock In {\em Proceedings of the IEEE/CVF International Conference on
  Computer Vision (ICCV)}, pages 5148--5157.

\bibitem[Kingma and Ba, 2015]{KingBa15}
Kingma, D. and Ba, J. (2015).
\newblock Adam: A method for stochastic optimization.
\newblock In {\em International Conference on Learning Representations (ICLR)}.

\bibitem[Lee et~al., 2020]{lee2020coupled}
Lee, A., Singh, S.~S., and Vihola, M. (2020).
\newblock Coupled conditional backward sampling particle filter.
\newblock {\em Annals of Statistics}, 48(5):3066--3089.

\bibitem[L\'{e}onard, 2014]{Christian2014schrodinger}
L\'{e}onard, C. (2014).
\newblock A survey of the {S}chr\"{o}dinger problem and some of its connections
  with optimal transport.
\newblock {\em Discrete and Continuous Dynamical Systems}, 34(4):1533--1574.

\bibitem[Liu et~al., 2015]{liu2015faceattributes}
Liu, Z., Luo, P., Wang, X., and Tang, X. (2015).
\newblock Deep learning face attributes in the wild.
\newblock In {\em Proceedings of International Conference on Computer Vision
  (ICCV)}, pages 3730--3738.

\bibitem[Loshchilov and Hutter, 2017]{loshchilov2017sgdr}
Loshchilov, I. and Hutter, F. (2017).
\newblock {SGDR}: Stochastic gradient descent with warm restarts.
\newblock In {\em International Conference on Learning Representations}.

\bibitem[Luo et~al., 2023]{Luo2023IRSDE}
Luo, Z., Gustafsson, F.~K., Zhao, Z., Sj\"{o}lund, J., and Sch\"{o}n, T.~B.
  (2023).
\newblock Image restoration with mean-reverting stochastic differential
  equations.
\newblock In {\em Proceedings of the 40th International Conference on Machine
  Learning}, volume 202, pages 23045--23066. PMLR.

\bibitem[Mardani et~al., 2024]{mardani2024a}
Mardani, M., Song, J., Kautz, J., and Vahdat, A. (2024).
\newblock A variational perspective on solving inverse problems with diffusion
  models.
\newblock In {\em The Twelfth International Conference on Learning
  Representations}.

\bibitem[Martin et~al., 2023]{martin2023computing}
Martin, G.~M., Frazier, D.~T., and Robert, C.~P. (2023).
\newblock Computing {B}ayes: from then ‘til now.
\newblock {\em Statistical Science}, 39(1):1--17.

\bibitem[Meyn and Tweedie, 2009]{Meyn2009}
Meyn, S.~P. and Tweedie, R.~L. (2009).
\newblock {\em Markov chains and stochastic stability}.
\newblock Cambridge University Press, 2nd edition.

\bibitem[Rogers and Williams, 2000]{Rogers2000diffusion}
Rogers, L. C.~G. and Williams, D. (2000).
\newblock {\em Diffusions, Markov Processes, and Martingales}.
\newblock Cambridge University Press, 2nd edition.

\bibitem[S{\"a}rkk{\"a} and Svensson, 2023]{sarkka2023bayesian}
S{\"a}rkk{\"a}, S. and Svensson, L. (2023).
\newblock {\em Bayesian filtering and smoothing}, volume~17.
\newblock Cambridge university press.

\bibitem[Shi et~al., 2023]{shi2023diffusion}
Shi, Y., Bortoli, V.~D., Campbell, A., and Doucet, A. (2023).
\newblock Diffusion {S}chr\"odinger bridge matching.
\newblock In {\em Advances in Neural Information Processing Systems},
  volume~36.

\bibitem[Shi et~al., 2022]{Shi2022conditional}
Shi, Y., De~Bortoli, V., Deligiannidis, G., and Doucet, A. (2022).
\newblock Conditional simulation using diffusion {S}chr{\"o}dinger bridges.
\newblock In {\em Proceedings of the 38th Conference on Uncertainty in
  Artificial Intelligence}, volume 180, pages 1792--1802. PMLR.

\bibitem[Song et~al., 2020]{song2020denoising}
Song, J., Meng, C., and Ermon, S. (2020).
\newblock Denoising diffusion implicit models.
\newblock In {\em International Conference on Learning Representations}.

\bibitem[Song et~al., 2023]{song2023pseudoinverseguided}
Song, J., Vahdat, A., Mardani, M., and Kautz, J. (2023).
\newblock Pseudoinverse-guided diffusion models for inverse problems.
\newblock In {\em International Conference on Learning Representations}.

\bibitem[Song and Ermon, 2020]{Song2020improve}
Song, Y. and Ermon, S. (2020).
\newblock Improved techniques for training score-based generative models.
\newblock In {\em Advances in Neural Information Processing Systems},
  volume~33, pages 12438--12448.

\bibitem[Song et~al., 2021]{song2021scorebased}
Song, Y., Sohl-Dickstein, J., Kingma, D.~P., Kumar, A., Ermon, S., and Poole,
  B. (2021).
\newblock Score-based generative modeling through stochastic differential
  equations.
\newblock In {\em Proceedings of the 9th International Conference on Learning
  Representations}.

\bibitem[Tjelmeland, 2004]{tjelmeland2004using}
Tjelmeland, H. (2004).
\newblock Using all {M}etropolis--{H}astings proposals to estimate mean values.
\newblock preprint 4/2004, Norwegian University of Science and Technology,
  Trondheim, Norway.

\bibitem[Trippe et~al., 2023]{trippe2023diffusion}
Trippe, B.~L., Yim, J., Tischer, D., Baker, D., Broderick, T., Barzilay, R.,
  and Jaakkola, T.~S. (2023).
\newblock Diffusion probabilistic modeling of protein backbones in {3D} for the
  motif-scaffolding problem.
\newblock In {\em Proceedings of The 11th International Conference on Learning
  Representations}.

\bibitem[Uhlenbeck and Ornstein, 1930]{Uhlenbeck1930brownian}
Uhlenbeck, G.~E. and Ornstein, L.~S. (1930).
\newblock On the theory of the {B}rownian motion.
\newblock {\em Physical Review}, 36(5):823--841.

\bibitem[Vargas et~al., 2023]{Vargas2023denoising}
Vargas, F., Grathwohl, W., and Doucet, A. (2023).
\newblock Denoising diffusion samplers.
\newblock In {\em Proceedings of the 11th International Conference on Learning
  Representations}.

\bibitem[Wang et~al., 2023]{wang2023exploring}
Wang, J., Chan, K.~C., and Loy, C.~C. (2023).
\newblock Exploring {CLIP} for assessing the look and feel of images.
\newblock In {\em Proceedings of the AAAI Conference on Artificial
  Intelligence}, volume~37, pages 2555--2563.

\bibitem[Whiteley and Lee, 2014]{Whiteley2014twistedPF}
Whiteley, N. and Lee, A. (2014).
\newblock Twisted particle filters.
\newblock {\em The Annals of Statistics}, 42(1):115 -- 141.

\bibitem[Wu et~al., 2023]{wu2023practical}
Wu, L., Trippe, B., Naesseth, C., Blei, D., and Cunningham, J.~P. (2023).
\newblock Practical and asymptotically exact conditional sampling in diffusion
  models.
\newblock In {\em Advances in Neural Information Processing Systems},
  volume~36, pages 31372--31403. Curran Associates, Inc.

\bibitem[Zhang et~al., 2018]{zhang2018cvpr}
Zhang, R., Isola, P., Efros, A.~A., Shechtman, E., and Wang, O. (2018).
\newblock The unreasonable effectiveness of deep features as a perceptual
  metric.
\newblock In {\em Proceedings of the IEEE Conference on Computer Vision and
  Pattern Recognition (CVPR)}, pages 586--595.

\bibitem[Zhao et~al., 2024]{Zhao2024rsta}
Zhao, Z., Luo, Z., Sj\"{o}lund, J., and Sch\"{o}n, T.~B. (2024).
\newblock Conditional sampling within generative diffusion models.
\newblock {\em arXiv preprint arXiv:2409.09650}.

\end{thebibliography}
\bibliographystyle{apalike}

%%%%%%%%%%%%%%%%%%%%%%%%%%%%%%%%%%%%%%%%%%%%%%%%%%%%%%%%%%%%
\section*{Checklist}

 \begin{enumerate}

 \item For all models and algorithms presented, check if you include:
 \begin{enumerate}
   \item A clear description of the mathematical setting, assumptions, algorithm, and/or model. [Yes]
   \item An analysis of the properties and complexity (time, space, sample size) of any algorithm. [Yes]
   \item (Optional) Anonymized source code, with specification of all dependencies, including external libraries. [Yes]
 \end{enumerate}

 \item For any theoretical claim, check if you include:
 \begin{enumerate}
   \item Statements of the full set of assumptions of all theoretical results. [Yes]
   \item Complete proofs of all theoretical results. [Yes]
   \item Clear explanations of any assumptions. [Yes]     
 \end{enumerate}

 \item For all figures and tables that present empirical results, check if you include:
 \begin{enumerate}
   \item The code, data, and instructions needed to reproduce the main experimental results (either in the supplemental material or as a URL). [Yes]
   \item All the training details (e.g., data splits, hyperparameters, how they were chosen). [Yes]
         \item A clear definition of the specific measure or statistics and error bars (e.g., with respect to the random seed after running experiments multiple times). [Yes]
         \item A description of the computing infrastructure used. (e.g., type of GPUs, internal cluster, or cloud provider). [Yes]
 \end{enumerate}

 \item If you are using existing assets (e.g., code, data, models) or curating/releasing new assets, check if you include:
 \begin{enumerate}
   \item Citations of the creator If your work uses existing assets. [Yes]
   \item The license information of the assets, if applicable. [Not Applicable]
   \item New assets either in the supplemental material or as a URL, if applicable. [Not Applicable]
   \item Information about consent from data providers/curators. [Not Applicable]
   \item Discussion of sensible content if applicable, e.g., personally identifiable information or offensive content. [Not Applicable]
 \end{enumerate}

 \item If you used crowdsourcing or conducted research with human subjects, check if you include:
 \begin{enumerate}
   \item The full text of instructions given to participants and screenshots. [Not Applicable]
   \item Descriptions of potential participant risks, with links to Institutional Review Board (IRB) approvals if applicable. [Not Applicable]
   \item The estimated hourly wage paid to participants and the total amount spent on participant compensation. [Not Applicable]
 \end{enumerate}

 \end{enumerate}

\clearpage
\onecolumn
\appendix
\section{Additional background}\label{app:background}

In this section we present additional background on generative diffusions in general. In particular, we show the foundation of sampling within generative diffusions, and how we can further leverage generative diffusions for conditional sampling.

\subsection{Unconditional generative models}\label{subsec:unconditional-generative-models}
There are two popular classes of diffusion-based generative models in the community, that are, denoising diffusion models, and Schr\"{o}dinger bridges. 

\subsubsection{Denoising diffusion models}\label{subsubsec:denoising-diffusion-models}
Denoising diffusion models~\citep{song2021scorebased} have recently received a lot of attention as a general-purpose sampler for generative modelling~\citep{song2021scorebased} or statistical inference~\citep{Vargas2023denoising}.
At the core, given a target distribution $\pi_0$ and a time-homogeneous stochastic differential equation (SDE)
\begin{equation}
    \label{eq:fwd-sde-base-background}
    \dd X_t = f(X_t) \dd t + \dd W_t, \qquad  X_0 \sim \pi_0,
\end{equation}
with stationary distribution $\pi_{\mrm{ref}}$, they sample from $\pi_0$ by ``denoising''~\eqref{eq:fwd-sde-base-background} by means of Doob's $h$-transform~\citep{Rogers2000diffusion}.
Formally, let $\pi_t$ be the distribution of $X_t$ under~\eqref{eq:fwd-sde-base-background}, and assume that we know how to sample from $\pi_T$ for some $T > 0$.
We can obtain approximate samples from $\pi_0$ by sampling from
\begin{equation}
    \label{eq:bwd-sde-base-background}
    \begin{split}
        \dd U_t &= f_{\mrm{rev}}(U_t, t) \dd t + \dd B_t,\\
        U_0 & \sim \pi_T,
    \end{split}
\end{equation}
where $B$ is another Brownian motion, and we write $f_{\mrm{rev}}(U_t, t) \coloneqq -f(U_t)  + \nabla \log \pi_{T-t}(U_t)$.

When $T \gg 1$, and under ergodicity guarantees~\citep{Meyn2009}, $\pi_T \approx \pi_{\mrm{ref}}$, so that if~\eqref{eq:fwd-sde-base-background} is chosen such that $\pi_{\mrm{ref}}$ is easy to sample from, then the only remaining blocker is computing the score $\nabla \log \pi_{T-t}$.
However, obtaining an expression for $\pi_t$ (and therefore for $\nabla \log \pi_{T-t}$) is in general not possible, and denoising diffusion models usually rely on score matching~\citep{hyvarinen2005score}, whereby an approximation $s_{\theta^*}(t, x_t)$ to $\nabla \log \pi_{t}$ is learnt under samples by minimising the loss function
\begin{equation}
    \label{eq:score-loss}
    \calL(\theta) = \int_{0}^T \bbE\left[\norm{s_{\theta}(X_t, t) - \nabla \log \pi_t(X_t \mid X_0)}^2 \dd t\right],
\end{equation}
with the expectation being taken under $\bbP$ in~\eqref{eq:fwd-sde-base-background}.
In practice, the method is implemented in discrete time, SDEs are sampled using numerical integrators, and the loss~\eqref{eq:score-loss} becomes
\begin{equation}
    \label{eq:score-loss-discrete-background}
    \begin{split}
        \calL(\theta) &=\sum_{k=1}^K \bbE\left[\norm{s_{\theta}(X_{t_k}, t_k) - \nabla \log \pi_{t_k}(X_{t_k} \mid X_0)}^2\right] \\
        &=\sum_{k=1}^K \bbE\left[\norm{s_{\theta}(X_{t_k}, t_k) - \nabla \log \pi_{t_k}(X_{t_k} \mid X_{t_{k-1}})}^2\right],
    \end{split}
\end{equation}
with $0 = t_0 < t_1 < \cdots < t_K = T$ being a given integration grid used also at evaluation time to simulate from~\eqref{eq:bwd-sde-base-background}.
When~\eqref{eq:fwd-sde-base-background} is chosen, as is often the case, to be linear~\citep[for example, as an Ornstein--Uhlenbeck process,][]{Uhlenbeck1930brownian}, $\pi_t(\cdot \mid X_0)$ can be computed and sampled from exactly so that the first version of~\eqref{eq:score-loss-discrete-background} incurs no error other than the sampling noise.
When it is not, as in the next section, the second version is more amenable to computations (and is also more numerically stable), at the cost of storing trajectories under $\bbP$.

\subsubsection{Schrödinger bridges}\label{subsubsec:schrodinger-bridges}
A strong drawback of denoising diffusion models lies in the assumption $T \gg 1$, under which $X_T$ is only approximately distributed according to $\pi_{\mrm{ref}}$.
Under this assumption, the denoising process~\eqref{eq:bwd-sde-base-background} will need to be numerically simulated for a large number of steps $K$, resulting in an inefficient and expensive procedure for both training and testing procedures.
To mitigate this, it is preferable to explicitly bridge between the distributions $\pi_0$ and $\pi_{\mrm{ref}}$ in a finite-time horizon, that is, find a forward SDE~\eqref{eq:fwd-sde-base-background} and a time reversal~\eqref{eq:bwd-sde-base-background} such that $\pi_T$ is exactly $\pi_{\mrm{ref}}$ for a finite $T$.
There are infinitely many such bridges, and in practice, we resort to the one that is in some sense easy to simulate.

Schrödinger bridges~\citep{Christian2014schrodinger,Chen2021optimaltransport} constitute a class of bridges which aim at minimising the total energy expense to move from $\pi_0$ to $\pi_T$ and vice versa.
Given a reference path measure $\bbQ$ on the set of continuous paths $\calC([0,T], \bbR^d) \eqqcolon \calC$, the Schrödinger bridge solution $\bbP^{\mrm{SB}}$ is defined as the (unique) solution to
\begin{equation}
    \label{eq:kl-schrodinger}
    \bbP^{\mrm{SB}} = \argmin_{\bbP \in \calP(\calC)}
    \left\{ \kl{\bbP}{\bbQ} \;\big|\; \bbP_0 = \pi_0, \bbP_T = \pi_T \right\},
\end{equation}
where $\calP(\calC)$ denotes all the path measures on $\calC$.
In general, solving~\eqref{eq:kl-schrodinger} directly is not possible, and instead, one needs to rely on alternative ways of solving the Schrödinger bridge.
The two main methods to do so are given by iterative proportional fitting~\citep[IPF,][]{deBortoli2021diffusion} and Schrödinger bridge matching~\citep[SBM,][]{shi2023diffusion} which we review below.

The IPF method follows from the fact that the solution of~\eqref{eq:kl-schrodinger} is given by the fixed point of a sequence of time-reversal operations: starting from $\bbP^0 = \pi_0\, \bbQ_{|0}$, one constructs a sequence $\bbP^{2n + 1} = \pi_T \{\iota \bbP^{2n}\}_{|T}$, $\bbP^{2n + 2} = \pi_0 \{\iota \bbP^{2n + 1}\}_{\mid T}$ by iteratively learning the time-reversal $\iota \bbP^n$ of the previous bridge $\bbP^n$, where we denote $\iota$ as a time-reversal operation.

The SBM method is given by the characterisation of $\bbP^{\mrm{SB}}$ as the unique measure in the set of Markov processes $\calM$ which also belongs to the reciprocal class of $\bbQ$, i.e., $\calR(\bbQ) \coloneqq \left\{\bbP \in \calP(\calC) \mid \bbP = \bbP_{0,T} \bbQ_{\mid 0, T}\right\}$ defined as the set of measures that correspond to bridges under the dynamics of $\bbQ$.
To find the Schrödinger bridge $\bbP^{\mrm{SB}}$, one can then construct a sequence of measures in $\calP(\calC)$ as follows: $\bbP^0 = (\pi_0 \otimes \pi_T) \bbQ_{\mid 0, T} \in \calR(\bbQ)$, and then iteratively project the approximate measure onto $\calM$ and $\calR(\bbQ)$.
Formally,
\begin{equation}
    \label{eq:bridge-proj}
    \begin{split}
        \bbP^{2n + 1} &= \argmin_{\bbP \in \calM} \left\{ \kl{\bbP^{2n}}{\bbP}\right\}, \\
        \bbP^{2n + 2} &= \bbP^{2n + 1}_{0,T} \bbQ_{\mid 0, T}.
    \end{split}
\end{equation}
Given $\bbP^{2n + 1}$, sampling from $\bbP^{2n + 2}$ can be done by first sampling from $\bbP^{2n + 1}_{0, T}$, discarding intermediate steps, and then sampling from the diffusion bridge $\bbQ_{\mid 0, T}$ by, for instance, Doob's $h$-transform.
The first step can then be done under samples from $\bbP^{2n + 1}$ by techniques akin to score matching (see, Section~\ref{subsubsec:denoising-diffusion-models}).
Under weak conditions, it can be shown that the sequence $\bbP^n$ converges to $\bbP^{\mrm{SB}}$ as $n\to\infty$, and that it preserves the marginal distributions $\bbP^n_0 = \pi_0$, $\bbP^n_T = \pi_T$ which is the key difference compared to the IPF approach.

\subsection{Conditional generative models}\label{subsec:conditional-generative-models}
In many applications, rather than unconditional sampling from a distribution $\pi(x)$, we are interested in generating conditional samples $\pi(x \mid y)$ for an observation $y$ under a joint model $\pi(x, y)$. 
In this section, we review several conditioning methods for diffusion models, focusing on bridging. 

\subsubsection{Conditional score matching}\label{subsubsec:conditional-diffusion-models}
Score-based generative models can readily generate conditional samples from pre-trained unconditional models by plugging $\nabla_x \log\pi(y \mid x)$ into the drift of~\eqref{eq:bwd-sde-base-background}.
However, it is in general hard to know $\pi(y \mid x)$ or its score, for example, in image restoration/classification applications~\citep{song2021scorebased}.
Due to this, one often needs to approximate a separate time-varying model for $\pi_t(y \mid x_t)$ or make domain-based heuristics, but this inevitably incurs additional errors for the conditional sampling.

\subsubsection{Conditional Schrödinger bridges}\label{subsubsec:conditional-schrodinger-bridges}
\citet{song2021scorebased} which requires learning or approximating the likelihood $\pi_t(y \mid x_t)$.
By contrast, \citet{Shi2022conditional} propose a method to learn a conditional sampler only requiring being able to sample artificial observations $\pi(y \mid x)$.
The idea is built upon constructing a Schrödinger bridge (see Section~\ref{subsubsec:schrodinger-bridges}) for the augmented distribution $\pi(x, y)$, namely
\begin{equation}
    \label{eq:cond-schrodinger}
    \bbP^{\mrm{CSB}} = \argmin_{\substack{\bbP \in \calP(\calC) \\ \bbP_0 = \pi(y \mid x) \pi(x) \\ \bbP_T = \pi_{\mrm{ref}}(x \mid y) \, \pi(y)}}
    \kl{\bbP}{\bbQ}
\end{equation}
with reference measure $\bbQ$ given by the system of SDEs
\begin{equation}
    \label{eq:cond-schrodinger-sde}
    \begin{split}
        \dd X_t &= a(X_t) \dd t + \dd W_t,\quad \dd Y_t = 0,
    \end{split}
\end{equation}
with initial distribution $X_0, Y_0 \sim \pi(x, y)$ and an arbitrary drift $a$.
In Equation~\eqref{eq:cond-schrodinger}, $\pi_{\mrm{ref}}(x \mid y)$ refers to any informative approximation to $\pi(x\mid y)$ that is easy to sample from.
We can then apply the methodology introduced in Section~\ref{subsubsec:schrodinger-bridges} to solve Equation~\eqref{eq:cond-schrodinger}.

Importantly, under~\eqref{eq:cond-schrodinger} one samples $\pi(y) = \int \pi(y, \dd x)$ by repeatedly sampling a data-point $X \sim \pi(x)$ and then generating $Y \sim \pi(y \mid X)$.
This makes CSB ill-suited to problems for which one has only access to joint samples $(X, Y)$ from the data, rather than to an observation-generator. 
Furthermore, CSB requires \emph{training} a specialised conditional model, adding to the complexity and the approximate nature of the resulting conditional sampler.

\subsection{Twisted diffusion samplers}\label{subsubsec:twisted-diffusion-models}

When an observation model $p(y \mid x)$ is available, and given a discretised version of~\eqref{eq:denoising-diffusion}, it is possible to form an augmented model
\begin{equation}
    \label{eq:augmented}
    \begin{split}
        U_{t_{k+1}} &= U_{t_k} + f_{\mrm{rev}}(U_{t_k}, t_k) \, (t_k - t_{k-1}) + \epsilon_{k}, \quad U_0 \sim \pi_{\mrm{ref}}, \quad Y \sim \pi(y \mid U_T),
    \end{split}
\end{equation}
for which the conditional distribution of $U_T \mid y$ is $\pi(\cdot \mid y)$.
This corresponds to a state-space model with a final observation $Y = y$, which can therefore be sampled by running the reverse diffusion~\eqref{eq:augmented} and then correcting the samples via importance sampling at the final time step.
This is however well-known to be inefficient~\citep{Whiteley2014twistedPF} as the unconditional samples $U_T$ are unlikely to be distributed according to $\pi(\cdot \mid y_T)$.

In order to efficiently sample from $\pi(\cdot \mid y)$, \citet{wu2023practical} proposed to form the twisted model
\begin{equation}
    \label{eq:twisted}
        \tilde{U}_{t_{k+1}} \sim \tilde{p}(u_{t_{k+1}} \mid y, u_{t_k}), \qquad \tilde{U}_0 \sim \pi_{\mrm{ref}}
\end{equation}
enabling the following posterior distribution
\begin{equation}
    \label{eq:twisted-posterior}
        \pi(U_{t_{0:K}} \mid y)
            =\pi_{\mrm{ref}}(U_0) \prod_{k=0}^{K-1} \tilde{p}(u_{t_{k+1}} \mid y, u_{t_k}) \frac{p(y \mid u_{t_{k+1}}) p(u_{t_{k+1}} \mid u_{t_k})}{p(y \mid u_{t_{k+1}}) \tilde{p}(u_{t_{k+1}} \mid y, u_{t_k})},
\end{equation}
which marginally recovers $\pi(U_T \mid y)$.
Once an approximation $\tilde{p}(u_{t_{k+1}} \mid y, u_{t_k})$ to the true conditional process $p(u_{t_{k+1}} \mid y, u_{t_k})$ is chosen, the twisted model~\eqref{eq:twisted} can then be sampled from using particle filtering, with a proposal $\tilde{p}(u_{t_{k+1}} \mid y, u_{t_k})$ and the corresponding importance weights as given in~\eqref{eq:twisted-posterior}, see also~\citet{wu2023practical,Whiteley2014twistedPF} for more details.
While principled, this approach presents at least the following two issues: (i) it requires the likelihood model $\pi(y \mid x)$ to be known, which is not the case in most applications of interest, (ii) and it is computationally intensive as forming the approximation $\tilde{p}$ requires the denoising score function to be differentiated~\citep[Section 3.2]{wu2023practical}.

\subsection{Pathwise conditioning}\label{subsubsec:pathwise-conditioning}
The idea of conditioning on the path of a diffusion process has been explored in the context of diffusion models by~\citet{trippe2023diffusion} for scaffolding and by \citet[concurrent with our paper]{dou2024diffusion} for the slighly more general setting of inverse problems.
The idea is that, given an observation $Y = y$, and provided that the forward noising diffusion for $X_t, Y_t$ is separable, one can form a sequence of observations $Y_{0:K}$ and then sample from the conditional distribution $\pi(X_{0:K} \mid Y_{0:K})$ by running the reverse diffusion~\eqref{eq:denoising-diffusion} conditioned on $Y_{0:K}$ for instance by using a particle filter as in~\citet{wu2023practical}.
Beside the need for a separable noising SDE, the method is asymptotically consistent under three assumptions:
\begin{enumerate}
    \item $p(X_T \mid y_0) \approx \pi_{\mrm{ref}}(X_T)$, that is, the forward noising process forgets the initial condition $y_0$ fast enough\label{item:forgetting};
    \item the number of particles $N$ is large enough to ensure that the particle filter is sufficiently close to the true posterior\label{item:pf-consistency};
    \item the forward noising dynamics is separable.\label{item:pf-sep}
\end{enumerate}
Condition~\ref{item:forgetting} is hard to verify in practice, condition~\ref{item:pf-consistency} is computationally expensive to ensure, and condition~\ref{item:pf-sep} does not cover all useful models, therefore making the method impractical for many applications.
Furthermore, once the reverse diffusion has been run, the samples only constitute an approximation to the conditional distribution, for a single observation path $Y_{t_{0:K}}$. Therefore the method needs to be repeated for another observation path $Y_{0:K}'$ if one wants an independent sample from $\pi(X_{0:K} \mid Y_{0:K}')$ and all intermediary samples need to be discarded accordingly.
Nonetheless, as we explain in Section~\ref{sec:related}, this approach is a biased approximation of the PMMH algorithm we develop in Section~\ref{subsec:separable-case} as it ignores the acceptance probability for the new path $Y'_{0:K}$.

\section{Particle Markov chain Monte Carlo}\label{app:pmcmc}

Particle Markov chain Monte Carlo~\citep[PMCMC]{andrieu2010particle} is a class of Markov chain Monte Carlo (MCMC) algorithms targetting the posterior distribution of a state-space model $\pi(x_{0:T}) \propto p(x_0) \prod_{t=1}^T p(x_t \mid x_{t-1}) \prod_{t=0}^T p(y_t \mid x_t)$, where $y_{t}$ is a given set of observations.

Two different versions of PMCMC exist: the first one, known as the particle independent Metropolis--Hastings (PIMH) algorithm, is a pseudo-marginal MCMC~\citep{andrieu2009pseudomarginal} algorithm that uses a particle filter to both propose a full new trajectory, and estimate the likelihood of the observations given the state, and the second one, known as the conditional SMC (cSMC) algorithm, is a Gibbs-like ensemble sampler~\citep{tjelmeland2004using} updates a trajectory sequentially using a particle filter. 

Both algorithms can also be used to perform parameter inference in state-space models, by targeting the joint posterior distribution of the parameters and the states. 
The corresponding instances of the algorithms are known as the particle marginal Metropolis--Hastings (PMMH) algorithm and the particle Gibbs (pGibbs) algorithm, respectively.
This is particularly useful in our context, where the observation trajectory $y_{1:T}$, where $y_0$ is our data, plays the role of a parameter in the state-space model with prior $p(y_{1:T} \mid y_0)$ being given by the forward noising diffusion model~\eqref{eq:denoising-diffusion}.

Give the current state of the Markov chain $X^n_{0:T}, Y^n_{1:T}$, the PMMH algorithm proceeds as follows:
\begin{enumerate}
    \item Propose a new trajectory $y_{1:T} \sim q(y_{1:T} \mid y^n_{1:T})$; \label{item:propose}
    \item Sample a new trajectory $x_{0:T} \sim \pi(x_{0:T} \mid y_{1:T})$ and the corresponding normalising constant estimator $\hat{\pi}(y_{1:T})$ from a particle filter; \label{item:sample}
    \item Compute the acceptance probability of the new state and observations $(x_{0:T}, y_{1:T})$ as
    \begin{equation}
        \label{eq:acceptance}
        \alpha = \frac{\hat{\pi}(y_{1:T}) q(y^n_{1:T} \mid y_{1:T}) p(y_{1:T}\mid y_0)}{\hat{\pi}(y^n_{1:T}) q(y_{1:T} \mid y^n_{1:T}) p(y^n_{1:T}\mid y_0)},
    \end{equation}
    \label{item:acceptance}
    \item Accept the new pair $(x_{0:T}, y_{1:T})$ with probability $\min\left\{1, \alpha\right\}$.
\end{enumerate}
In step~\ref{item:sample}, the trajectory is obtained as a fixed-lag smoothing estimate of the state trajectory~\citep[Chap. 12]{chopin2020book} and in step~\ref{item:acceptance}, the normalising constant estimator $\hat{\pi}(y^n_{1:T})$ is the one that was computed when the current state $(x^n_{0:T}, y^n_{1:T})$ was accepted.

On the other hand, the pGibbs algorithm proceeds as follows:
\begin{enumerate}
    \item Sample a new trajectory $y^{n+1}_{1:T} \sim \pi(y_{1:T} \mid x^n_{0:T}, y_0)$; \label{item:propose-pgibbs}
    \item Sample a new trajectory $x^{n}_{0:T} \sim K(\cdot \mid y_{1:T}, x^{n}_{0:T})$ from the so-called conditional SMC kernel $K$, given in Algorithm~\ref{alg:csmc}. \label{item:sample-pgibbs}
\end{enumerate}

While the full description of cSMC is given in Algorithm~\ref{alg:csmc}, we provide a simplified version of the method in Algorithm~\ref{alg:csmc-single} that considers a single time step, thereby recovering algorithms from~\citet{tjelmeland2004using}.

\begin{algorithm}[H]
    \caption{Conditional Sequential Monte Carlo (cSMC) for a single time step}
    \label{alg:csmc-single}
    \begin{algorithmic}
        \INPUT{$x_{0}$, $y_0$}
        \OUTPUT{A new $x_{0}$}
        \FOR{$n=2, \ldots, N$}
        \STATE{Sample $x_{0}^{n} \sim p(x_0)$}
        \ENDFOR
        \STATE{Set $x_0^1 = x_0$}
        \FOR{$n=1, \ldots, N$}
        \STATE{Compute $w^n = p(y_0 \mid x_0^n)$}
        \ENDFOR
        \STATE{Sample $m \in \{1, \ldots, N\}$ with probabilities proportional to $w^{1:N}$}
        \STATE{Set $x_0 = x_0^m$}
    \end{algorithmic}
\end{algorithm}

In practice, PMMH and PGibbs have different benefits and drawbacks: PMMH allows to sample the parameter $y_{1:T}$ directly, but is hindered by the variance of the likelihood estimator, while PGibbs is more stable but requires to sample the parameter $y_{1:T}$ conditionally on the state $x_{0:T}$, which can be inefficient when the two are highly correlated.
Notwithstanding, CSMC is usually preferred in practice because, contrary to PMMH, it recovers a perfect Gibbs sampler when $N \to \infty$, while the acceptance rate of PMMH will usually not converge to one, even in the limit of infinite particles~\citep[Appendix]{andrieu2018uniform}.

\section{Conditional sampling under linear observation models}\label{app:linear-observation}
Concurrently to our work,~\citet{dou2024diffusion} have extended~\citet{trippe2023diffusion}'s work to the case of linear observation models, whereby they are given a diffusion model targeting the unconditional distribution $\pi(x)$, and a linear observation model $y = A x + c \xi$, where $A$ is a linear operator, $\xi$ is a standard Gaussian noise, and $c > 0$ is the observation noise scale. Then they aim to sample from the conditional distribution $\pi(x \mid y)$.

Assuming that the forward noising SDE and its reverse denoising 
\begin{equation}
    \label{eq:denoising-diffusion-linear}
    \begin{split}
        \dd X_t = -\frac{\sigma^2}{2} X_t \dd t + \sigma \dd W^X_t, \quad
        \dd U_t = \mu^U(U_t, t) \dd t + \sigma \dd W^U_t, 
    \end{split}
\end{equation}
are known\footnote{Here we have assumed that the noising SDE had constant diffusion for the sake of ease of exposition.}, they proceed to sample from $\pi(x \mid y)$ by forming an augmented noising diffusion model for the joint distribution of $x_{[0, T]}$ and $y_{(0, T]}$ conditioned on $y_0$:
\begin{equation}
    \label{eq:denoising-diffusion-linear-cond}
    \begin{split}
        \dd Y_t = -\frac{\sigma^2}{2} Y_t \dd t + \sigma A \dd W_t, \qquad
        \dd X_t = -\frac{\sigma^2}{2} X_t \dd t + \sigma \dd W_t,
    \end{split}
\end{equation}
for the same noise process $W_t$ for both $X$ and $Y$ and $(X_0, Y_0) \sim \pi(x) \mathcal{N}(y; A x, c^2I)$

They then show that the reverse pair of SDEs follows, after discretisation:
\begin{equation}
    (u_{0:K}, v_{0:K}) \sim \mathbb{B}(u_{0:K}) \prod_{k=0}^K \mathcal{N}(v_k; A u_k, c^2 \lambda_k I),
\end{equation}
where the variance factor $\lambda_k$ can be computed in closed form.
As a consequence, once $(Y_{k})_{k=0}^K$ has been obtained from \eqref{eq:denoising-diffusion-linear-cond},
they can obtain approximate samples for $U_{0:K} \mid \{V_{0:K}=Y_{K:0}\}$ by using a particle filter.
This approach is highly similar to \citet{trippe2023diffusion}: it is a biased pseudo-marginal algorithm, and suffers from the same bias as they do.
Consequently, we can correct their approach in exactly the same way as we have corrected the method of \citet{trippe2023diffusion} in this article:
rather than implementing an approximate sampler for $\mathbb{B}(u_{0:K} \mid V_{0:K})$, we can implement a Markov kernel keeping it invariant such as the CSMC or correct pseudo-marginal kernels we propose in this article.

\section{Correction of the measurement path}
\label{sec:app-gibbs-correction}
One minor caveat of our Gibbs-CSMC sampler is that the path $Y_{1:K}$ was not drawn (marginally) from the backward model $\bbB$ but from $\bbF$.
Specifically, since the learnt reversal is almost always an estimated one in practice, the path $V_{K:1} = Y_{1:K}$ drawn from the forward process $\bbB$ is not a sample of the reversal $\bbB$, 
resulting in a statistical bias. 
However, we show that (i) this bias can be fully corrected with an intermediary Metropolis--Hasting step, and that (ii) this bias empirically does not impact the results when ignored. 
First, we show how to remove the bias. 
Define $Y_{1:K}^*$ as the proposal path sampled from the forward process, and $Y_{1:K}^{j}$ as the previous Gibbs chain sample (the same goes for $X_{1:K}^*$ and $X_{1:K}^j$), we accept/reject the proposal based on the probability
\begin{equation}
    \min\Biggl(1, \frac{q(Y_{1:K}^* \cond X_{0:K}^*, y_0) \, p(Y_{1:K}^j, X_{1:K}^j \cond X_0^j, y_0)}{q(Y_{1:K}^j \cond X_{0:K}^j, y_0) \, p(Y_{1:K}^*, X_{1:K}^* \cond X_0^*, y_0)}\Biggr),
    \label{equ:gibbs-mh-correction}
\end{equation}
where $q$ is the (approximate) backward process density, and $p$ is the (exact) forward density. 
This kernel then targets the correct distribution with the approximate prior diffusion. 
Lastly, we perform a new experiment to verify the bias of the mismatch (when ignored) by extending the MNIST super-resolution experiment. 
We remark that this experiment is on a Schr\"{o}dinger bridge model, where the issue would be more pronounced since \emph{both} the forward and backward are learnt, and, therefore are more likely to be inconsistent. 
We use the identical experiment setting as in Section~\ref{sec:experiment-sb}, except that now we involve the MH kernel correction of Equation~\eqref{equ:gibbs-mh-correction} in the Gibbs-CSMC sampler (using 10 particles). 
The results are shown in Table~\ref{tbl:gibbs-mh-correction}, and we conclude that for this experiment the bias is not significant, and in practice, the MH correction may indeed be unnecessary.

\begin{table}[t!]
    \caption{The effect of correcting the measurement path in Section~\ref{sec:app-gibbs-correction}. 
    The numbers are averaged over 100 independent runs and 100 MC samples.
    We see that Gibbs-CSMC with correction has hardly differs from its uncorrected version, showing that the bias is empirically negligible.}
    \label{tbl:gibbs-mh-correction}
    \centering
    \begin{tabular}{@{}llll@{}}
    \toprule
    Method                        & PSNR (std)       & SSIM (std)      & MH acceptance ratio (std) \\ \midrule
    Gibbs-CSMC without correction & 10.6373 (2.0228) & 0.2806 (0.1368) & 1.00 (0.00)               \\
    Gibbs-CSMC with MH correction & 10.6370 (2.0099) & 0.2820 (0.1382) & 0.48 (0.50)               \\ \bottomrule
    \end{tabular}
\end{table}

\section{Experiment settings}
\label{appendix:experiment}
In this section, we detail our experimental setting and report additional results supporting the conclusions of our experiments. 

\subsection{Common settings}
\label{appendix:common}
Unless otherwise stated, our Gibbs-CSMC method applies the explicit backward sampling in Section~\ref{subsubsec:tractable-forwards-dynamics} and the killing conditional resampling~\citep{karppinen2023bridge}, and our PMCMC method applies the PCN proposal of~\citet{cotter2013crank} and stratified resampling~\citep[Chap. 9]{chopin2020book}.

For image-related experiments, we use a standard UNet commonly used in denoising diffusion models~\citep[see][]{Ho2020} with three downsampling and three upsampling layers, and we use the pixel shuffle method for upsampling in the UNet. 
The initial number of convolution features is 64, followed by 128, and 256 in the three layers. 
The detailed construction of the neural network is given in our companion code repository at \url{\codeaddress}.

We randomly split the MNIST dataset into 60,000 training and 10,000 test data points. 
As for the CelebaHQ dataset~\citep{liu2015faceattributes}, we split it into 29,000 training and 1,000 test data points. 
All the image data are normalised within the range $[0, 1]$.

All experiments are implemented in JAX~\citep{jax2018github} with the Flax~\citep{flax2020github} neural network backend. 
The experiments are conducted on an NVIDIA A100 80G GPU and the code required to reproduce our empirical findings can be found at \url{\codeaddress}.

\subsection{High-dimensional synthetic conditional sampling}
\label{appendix:gp}
In this experiment, our goal is to sample the posterior distribution of a Gaussian process model
\begin{equation*}
    \begin{split}
        f(\tau) &\sim \mathrm{GP}\bigl( 0, C(\tau, \tau') \bigr),\\
        y(\tau) \mid f(\tau) &\sim \mathrm{N}(0, \Xi),
    \end{split}
\end{equation*}
where $C$ is an exponential covariance function with length scale $\ell=1$ and magnitude $\sigma=1$, and the observation noise covariance $\Xi$ is a unit diagonal matrix. 
The GP regression is taken on 100 test points uniformly placed at $\tau \in [0, 5]$, resulting in a 100-dimensional target conditional distribution to sample from. 
The forward noising process is chosen to be $\dd X_t = -0.5 \, X_t \dd t + \dd W_t$ at $t\in[0, 1]$, so that we can compute the associated score function exactly by computing the marginal mean and covariance of $X_t$.
For each single experiment and each method, we generate 10,000 conditional samples, and then repeat the experiment independently 100 times to report the results. To show how the samples approximate the true distribution, we compute the KL divergence, Bures--Wasserstein distance, and the mean absolute errors on the marginal means and variances. 

While the CSGM method needs the likelihood distribution $p(y \mid x_t)$ to work~\citep{song2021scorebased}, in this Gaussian special case, the likelihood is analytically tractable, and hence the method can be applied exactly. 
For the twisted PF, we apply the GP observation model as the twisting function as per~\citet[Eq. (8)]{wu2023practical}. 

\begin{figure}
    \centering
    \includegraphics[width=.99\linewidth]{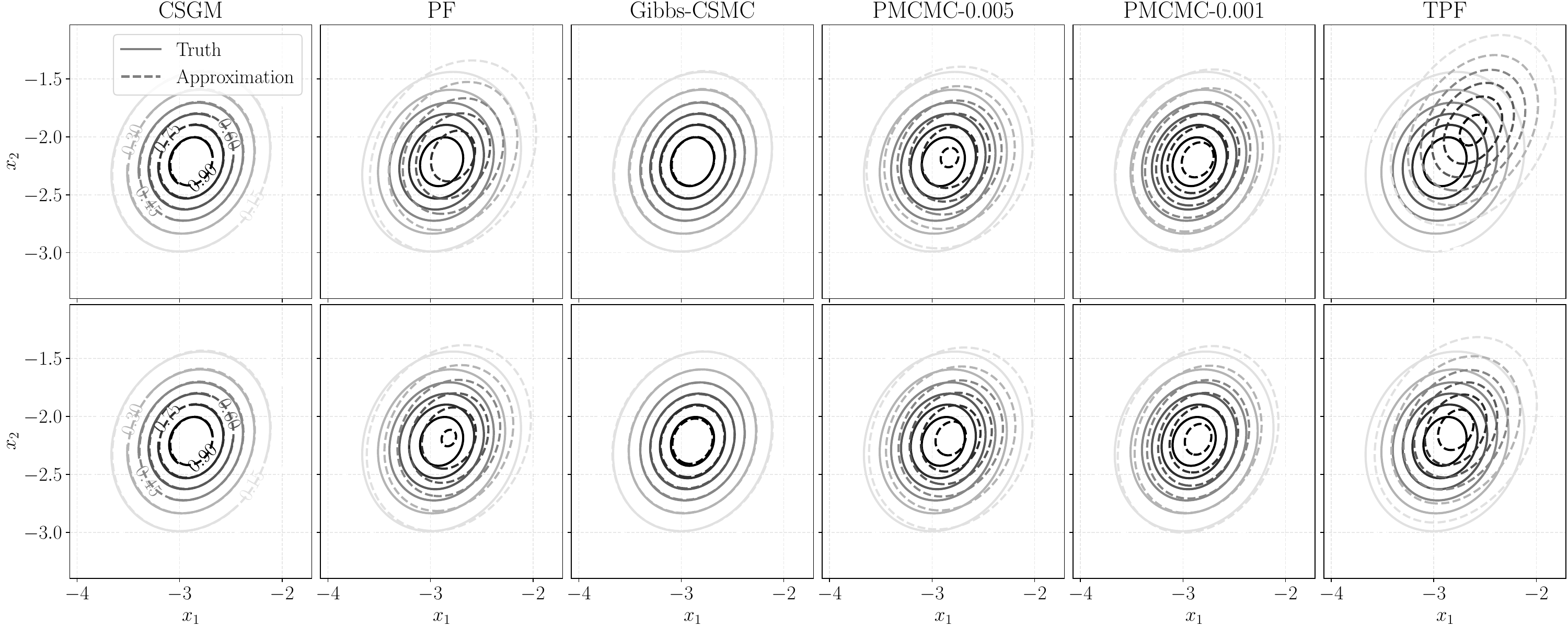}
    \caption{Contour plot of two-dimensional marginal densities for the true and approximate distributions in Section~\ref{sec:experiment-gp-regression}. The contour level lines are consistent in all the figures. The first and second rows show the results when using 10 and 100 particles, respectively. The figure shows that the Gibbs-CSMC method is visibly the best, and that it works well, even when using a small number of particles.}
    \label{fig:app-toy-contours}
\end{figure}

Figure~\ref{fig:app-toy-contours} compares the true and approximate distributions over two marginal random variables. 
The figure shows that the Gibbs-CSMC method outperforms others, followed by the PMCMC method. 
Increasing the number of particles gives noticeable improvements for the PF and TPF methods, meaning that these two methods require larger numbers of particles to work well for high-dimensional conditional samplings.

\begin{figure}[t!]
    \centering
    \includegraphics[width=.99\linewidth]{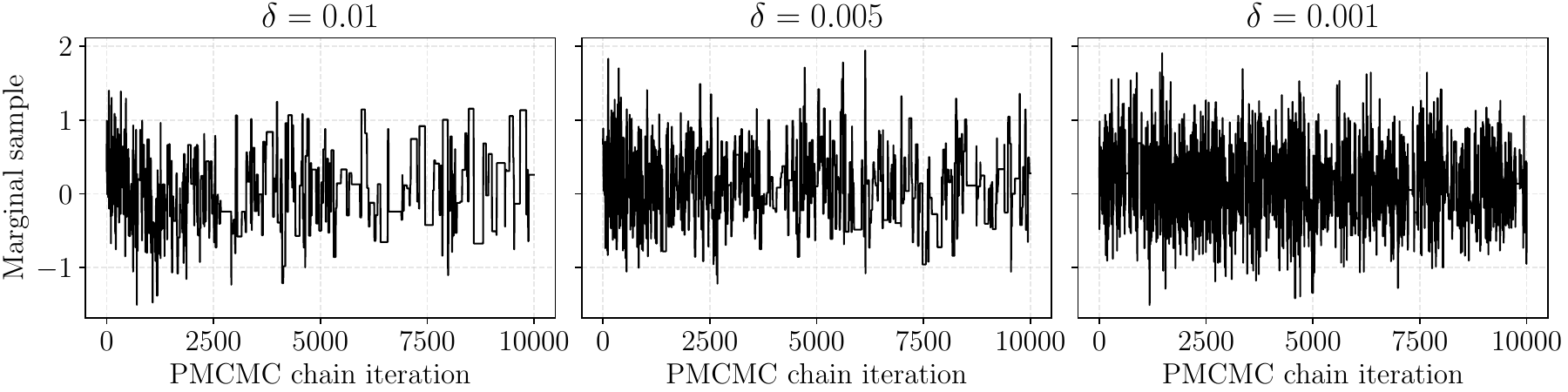}
    \caption{Trace plot of the PMCMC chain for the conditional sampling in Section~\ref{sec:experiment-gp-regression} when using 10 particles and different $\delta$. 
    The traces are downsampled by 2 for visibility. 
    We see that the calibration of $\delta$ reflects the effectiveness of the MCMC chain.}
    \label{fig:app-toy-pmcmc-trace}
\end{figure}

Figure~\ref{fig:app-toy-pmcmc-trace} shows the trace of the PMCMC chain of one marginal variable. 
We see that the parameter $\delta$ affects the autocorrelation of the MCMC chain, which is also reflected in Table~\ref{tbl:toy-errs} and Figure~\ref{fig:toy-autocorr}. 
The chain appears to be more effective when $\delta$ is small, meaning that the measurement path $t\mapsto Y_t$ in Section~\ref{subsec:separable-case} ideally should not move too much.
Furthermore, we also see that the chain seems to gradually degenerate as the chain iteration goes for larger moves in the $Y$ space.
This is a well-known defect of pseudo-marginal methods~\citep{andrieu2009pseudomarginal} in general, and comes from the fact that proposed paths are likely to be accepted originally due to the variance in the acceptance ratio, and then unlikely once the process has stabilised if the proposed path is too far from the current one.
Solving this problem is an interesting avenue of future work and likely involves redesigning the proposal kernel for the path $Y$ to take into account the specific structure of the model.

In Section~\ref{subsubsec:tractable-forwards-dynamics} we mentioned the memory cost due to storing the path if not implementing the online simulation of the bridge. 
In this experiment, we used 200 time steps, and the memory exploded when increased the number of steps to 20000 on a computer with 32GB RAM.

\subsection{Conditional sampling in non-separable noising processes}
\label{sec:app-sb}
In this section, we detail the experimental setting of Section~\ref{sec:experiment-sb}. 

To show the effect of using a non-separable noising process, we conduct quantitative and qualitative experiments using Schr\"{o}dinger bridges. 
In the quantitative experiment, we apply the same GP model in Appendix~\ref{appendix:gp}, but we scale the observation noise covariance to $0.1$ to increase the correlation between $X$ and $Y$. 
We use a Wishart distribution to randomly generate the reference covariance matrix $Q = q \otimes q$, where $\otimes$ is the outer product, and $q \sim \mathcal{N}(0, I_d)$.
We then consider a Gaussian Schr\"{o}dinger bridge between $\pi(x, y)$ and $\mathcal{N}(0, Q)$, with reference process $\dd Z_t = \dd W_t$ given as the standard Brownian motion,
the formulation of which is analytically tractable~\citep{Bunne2023}.

For the qualitative experiments on MNIST images, we train a Schr\"{o}dinger bridge with reference process $\dd Z_t = -0.5 \, \beta_t \, Z_t \dd t + \sqrt{\beta_t} \dd W_t$ at $t\in[0, 0.5]$, where $\beta_t \coloneqq (b_{\mathrm{max}} - b_{\mathrm{min}}) \, / \, (T - t_0) \, t + (b_{\mathrm{min}} \, T - b_{\mathrm{max}} \, t_0) \, / \, (T - t_0)$, and we set $b_{\mathrm{max}} = 5$, $b_{\mathrm{max}} = 0.02$, $t_0=0$, and $T=0.5$. 
The reference distribution $\pi_{\mathrm{ref}}$ at $T$ is a standard unit Gaussian. 
The number of Schr\"{o}dinger bridge iterations is 20. 
At each Schr\"{o}dinger bridge iteration, we run 20 epochs training the forward and backward models, with spatial batch size 64 and temporal batch size 32.
We use the Adam~\citep{KingBa15} optimiser with learning rate decaying from $2 \times 10^{-4}$ to $2 \times 10^{-6}$ in a cosine curve~\citep{loshchilov2017sgdr}.
For more details on the algorithm used to train the bridge, see~\citet[Algorithm 1]{deBortoli2021diffusion}.

\begin{figure}[t!]
    \centering
    \includegraphics[width=.8\linewidth]{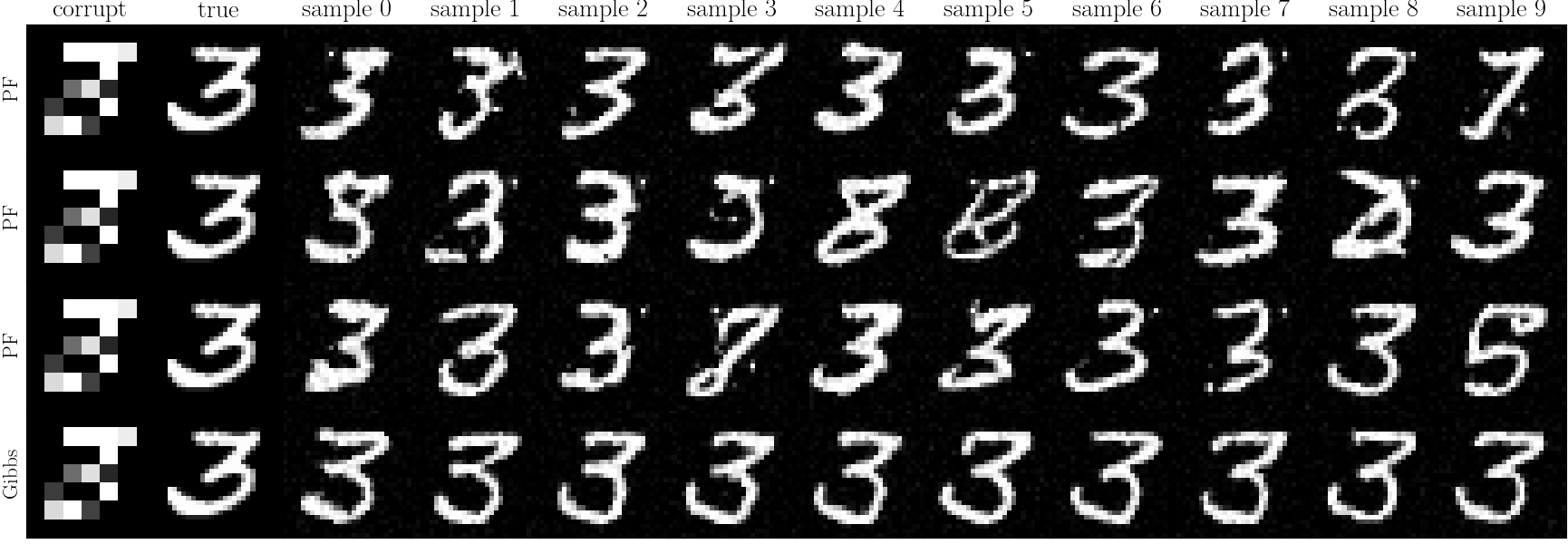}
    \includegraphics[width=.8\linewidth]{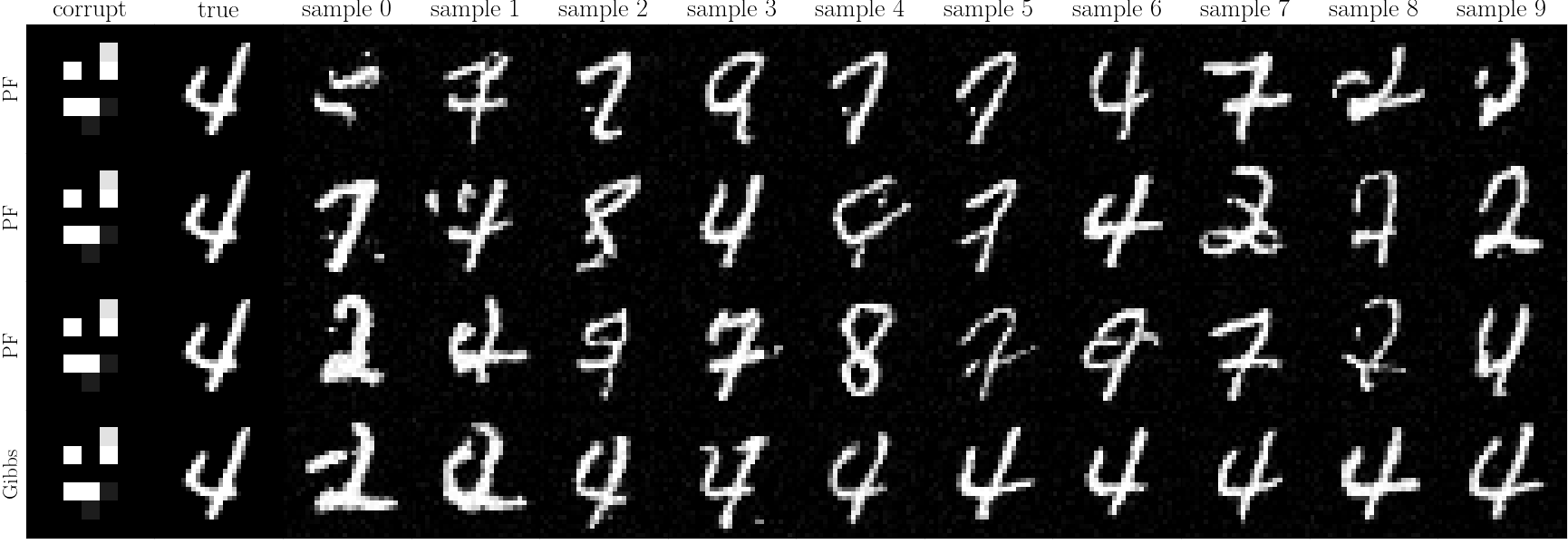}
    \includegraphics[width=.8\linewidth]{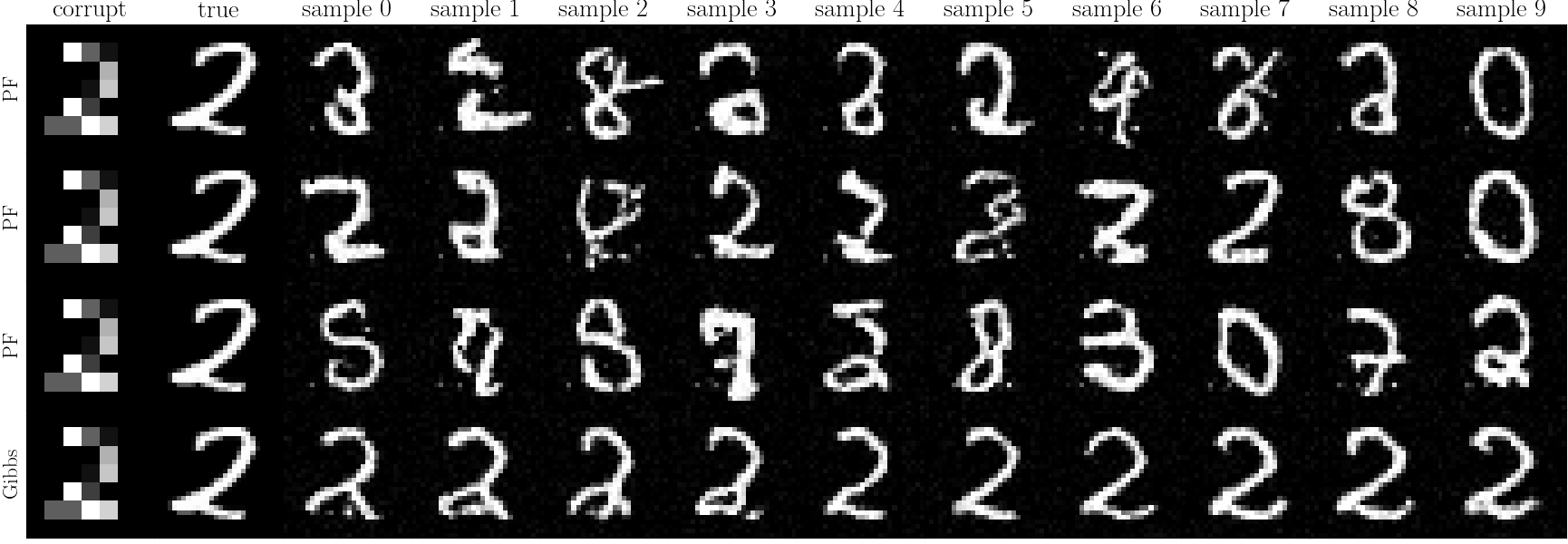}
    \includegraphics[width=.8\linewidth]{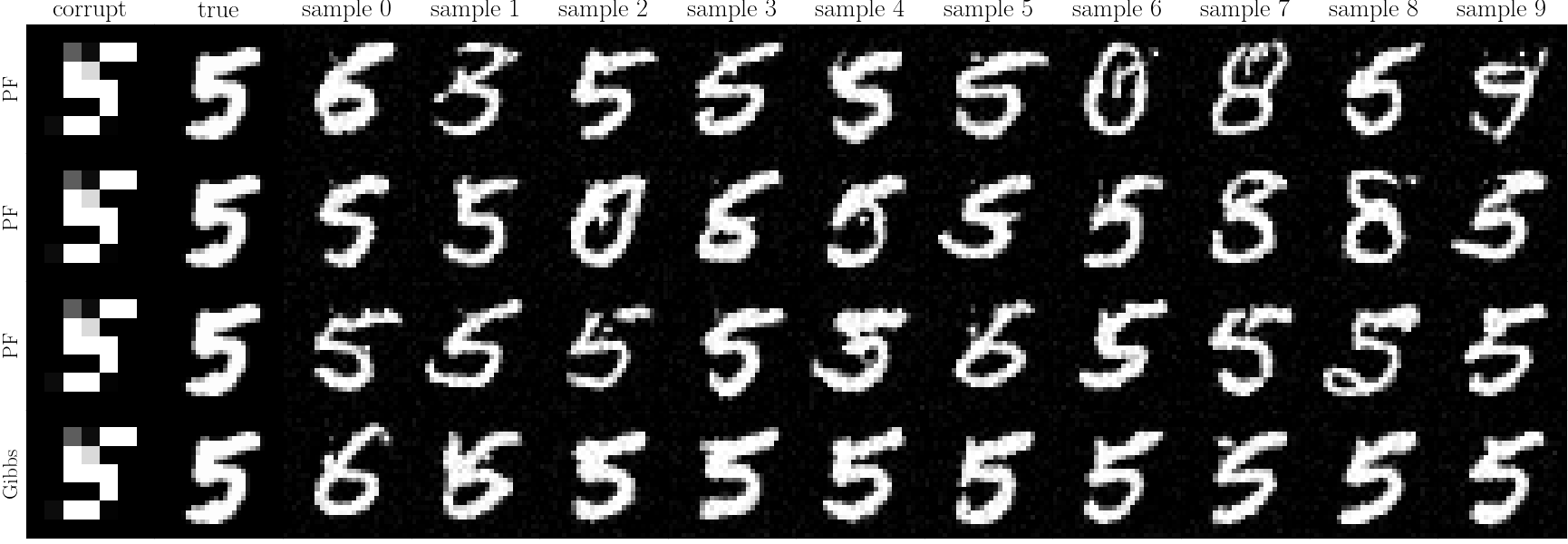}
    \caption{MNIST super-resolution (x4) on a Schr\"{o}dinger bridge. The four panels illustrate four test samples. 
    In each panel, the first to last rows show the samples of PF (using random intialisation), PF (using all-zeros initialisation), PF (using linear interpolation initialisation), and our Gibbs-CSMC method, respectively. 
    The initialisation in PF affects the quality, with the linear interpolation slightly better than other initialisations.
    Nonetheless, all PF methods generate worse samples compared to Gibbs-CSMC.
    For Gibbs-CSMC, while the initial sample is poor, it gradually improves and reaches the `true' stationary distribution $\pi(x \mid y)$ in a few iterations of the MCMC chain.}
    \label{fig:app-sb-imgs}
\end{figure}

Figure~\ref{fig:app-sb-imgs} shows additional examples of the MNIST super-resolution task using this Schr\"{o}dinger bridge unconditional sampler model.
In this figure, we reproduce the results returned by PF with various initialisations for $X_0$ and compare them to our Gibbs-CSMC method. 
Specifically, in PF we generate the initial value for $X_0$ using random uniform values (between 0 and 1), all zeros, and a linear interpolation based on $y$.
This starting value is then used within the learnt non-separable noising diffusion to generate the measurement path is then used for the backward filtering routine.
When using our CSMC-Gibbs method, we did not observe significant differences arising from the choice of the initial $X_0$, and therefore report the result using an initial $X_0 = 0$ only.

As shown in Figure~\ref{fig:app-sb-imgs}, the initialisation of $X_0$ indeed affects the quality of PF samples due to the non-separability of the Schr\"{o}dinger bridge, while our Gibbs-CSMC sampler returns valid samples.
This visible bias stems from the fact that, under non-separability, the choice of $X_0$ has a direct impact on the distribution of the path $Y_{1:K} \mid Y_0=y, X_0$, as conditional independence is then not verified anymore.
For the path $Y_{1:K} \mid \{Y_0=y, X_0\}$ to be marginally distributed according to the forward noising diffusion~\citep[as required by][]{trippe2023diffusion}, one would need the initial value of $X_0$ to be distributed according to $\pi(x \mid y)$, which is the problem we are trying to solve and is therefore not feasible in practice.
While it is reasonable to assume that approximate samples from $\pi(x \mid y)$ can be obtained via other methods, this would incur additional computation and still not eliminate the bias.
On the other hand, our Gibbs-CSMC method provides a natural way to produce samples that iteratively converge to $\pi(x \mid y)$, explaining its improved performance.
% Even when the initial sample $X_0$ is poor, the MCMC chain can effectively burn the poor sample into a correct sample. 

\subsection{Image inpainting and super-resolution}
\label{app:imgs-samples}
In this section, we describe the image inpainting and super-resolution experiments of Section~\ref{sec:experiment-imgs}. 
These two tasks can be solved by our conditional samplers without dedicated training. 
Namely, given a corrupted image, we can obtain clean images by using our samplers as they are, on pre-trained generative diffusion models. 
More precisely, for any image $Z \sim \pi(\dd z)$, we split it into the unobserved $X$ and observed $Y$ parts, which correspondingly separate the (backward) generative diffusion of~\eqref{eq:separable} for $\pi(\dd z)$. 
When the task is super-resolution, we model the observed low-resolution image as a subsampling of the original image.
We then model the image inpainting and super-resolution as sampling from the unobserved pixels conditioned on the observed pixels.

For both tasks, we use a time-varying linear SDE as the noising process as in~\eqref{eq:separable}. 
Specifically, the forward process is given $\dd Z_t = -0.5 \, \beta_t \, Z_t \dd t + \sqrt{\beta_t} \dd W_t$ at $t\in[0, 2]$, where $Z_t \coloneqq [ X_t, Y_t ]$, $\beta_t \coloneqq (b_{\mathrm{max}} - b_{\mathrm{min}}) \, / \, (T - t_0) \, t + (b_{\mathrm{min}} \, T - b_{\mathrm{max}} \, t_0) \, / \, (T - t_0)$, and we set $b_{\mathrm{max}} = 5$, $b_{\mathrm{max}} = 0.02$, $t_0=0$, and $T=2$.

We first pre-train the unconditional model to approximate the resulting score function using the standard denoising score matching method of~\citet{song2020denoising}. 
The optimiser we used for training is Adam with a decaying cosine learning rate schedule, i.e., the learning rate starts from $2 \times 10^{-4}$ and ends at $2 \times 10^{-6}$ following a cosine curve~\citep{loshchilov2017sgdr}. 
The spatial and temporal batch sizes are 256.
We also apply gradient clipping and exponential moving average (with 0.99 decay rate per two iterations) as suggested in~\citet{Song2020improve}. 
We run the training for 3,000 epochs and use the model obtained in the last epoch.

In the inpainting task, we apply rectangles of sizes 15 and 32 for MNIST and CelebAHQ, respectively, spawning at random locations (uniformly) in the image. 
As for super-resolution, we use rates 4 and 2 for MNIST and CelebAHQ. 
Precisely, rate 2 means that for an image of size 32 by 32, we partition it into 16 by 16 squares, where each square has 2 by 2 pixels; at each square, we randomly select one pixel for the low-resolution image.
For each test (inpainting/super-resolution) image, we generate 100 restored image samples, and report the statistics as in Tables~\ref{tbl:mnist} and~\ref{tbl:celeba}. 
In the table, the LPIPS score uses an Alex net~\citep{zhang2018cvpr}. 
The commonly used Fr\'{e}chet inception distance is not applied here, since it is not clear how to use it to benchmark MCMC samples~\citep[cf. the arguments in][]{Cardoso2024monte}. 

For the CSGM method, we implement the conditional drift as per~\citet[][Sec. I.2]{song2021scorebased}. 
As for the TPF, we also implement the same twisting function as in~\citet[][Sec. 3.3]{wu2023practical}.

\begin{figure}
    \centering
    \includegraphics[width=.8\linewidth]{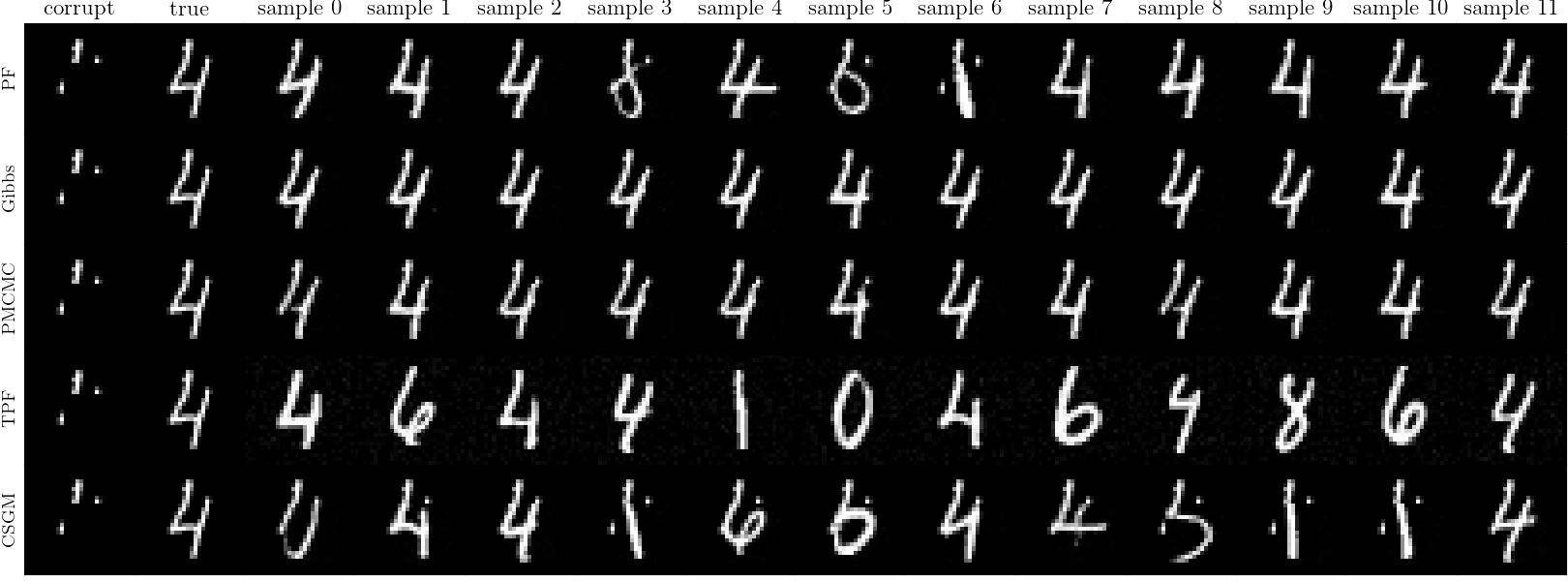}
    \includegraphics[width=.8\linewidth]{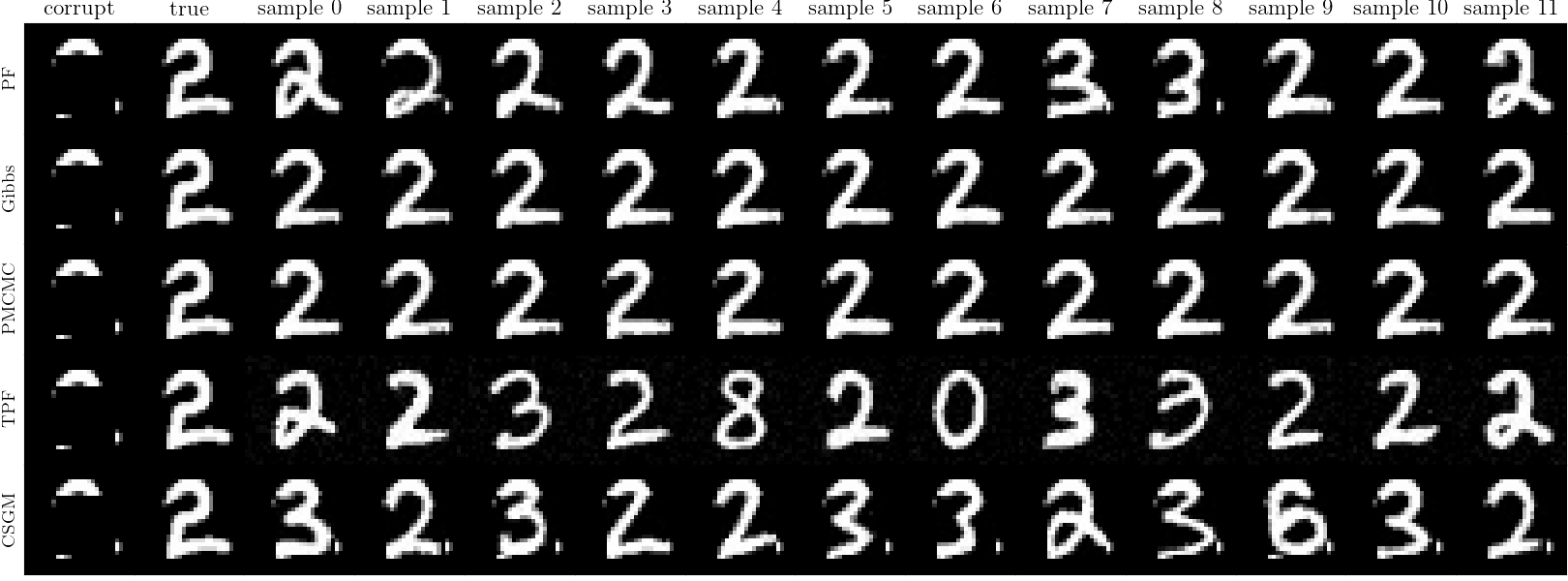}
    \includegraphics[width=.8\linewidth]{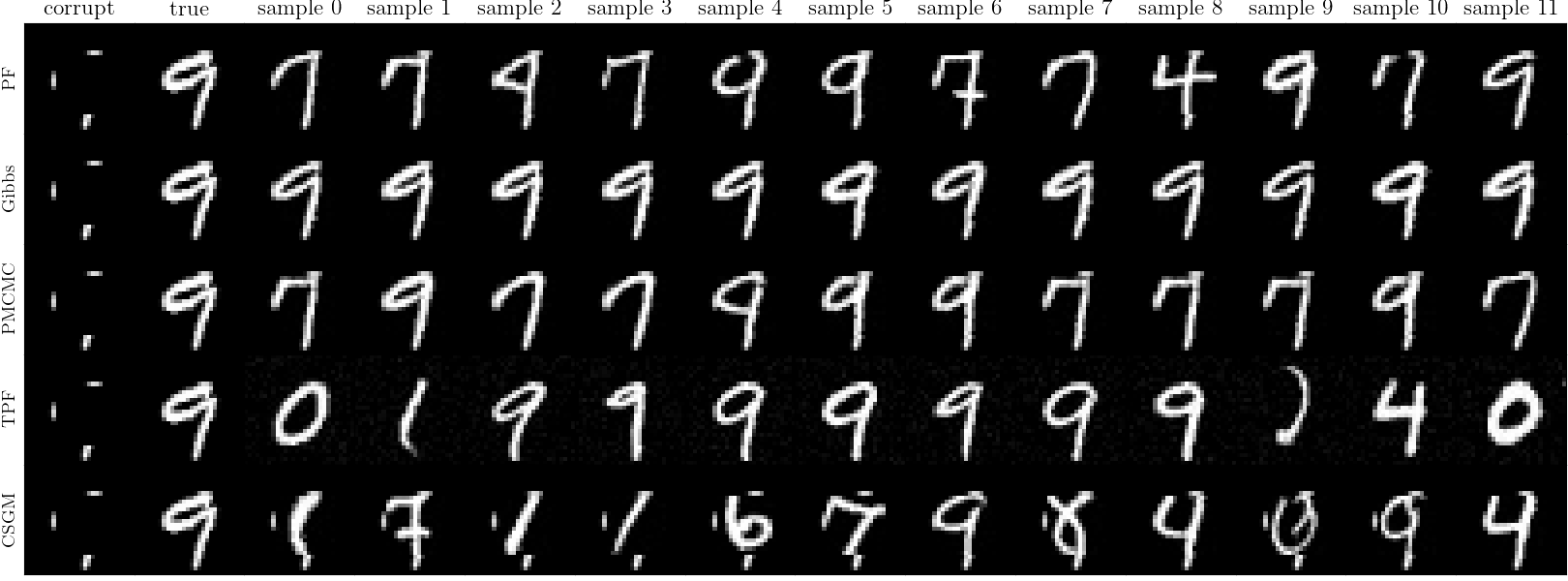}
    \includegraphics[width=.8\linewidth]{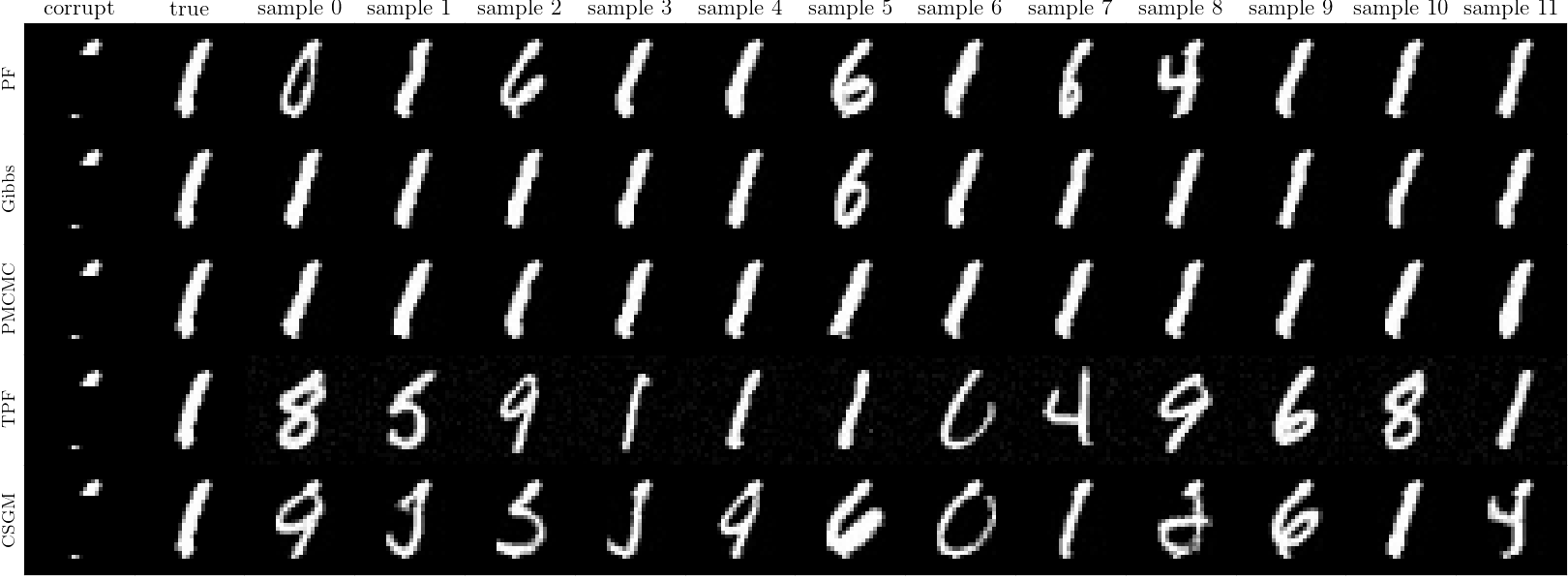}
    \caption{Examples of MNIST inpainting (using 100 particles) tasks. The figure has four panels, showing the results for four test images. In each panel, the first to last rows show the results of PF, Gibbs-CSMC, PMCMC-0.005, TPF, and CSGM, respectively. Wrong samples and artefacts appear less in our MCMC methods compared to others.}
    \label{fig:app-mnist-inpainting}
\end{figure}

\begin{figure}
    \centering
    \includegraphics[width=.8\linewidth]{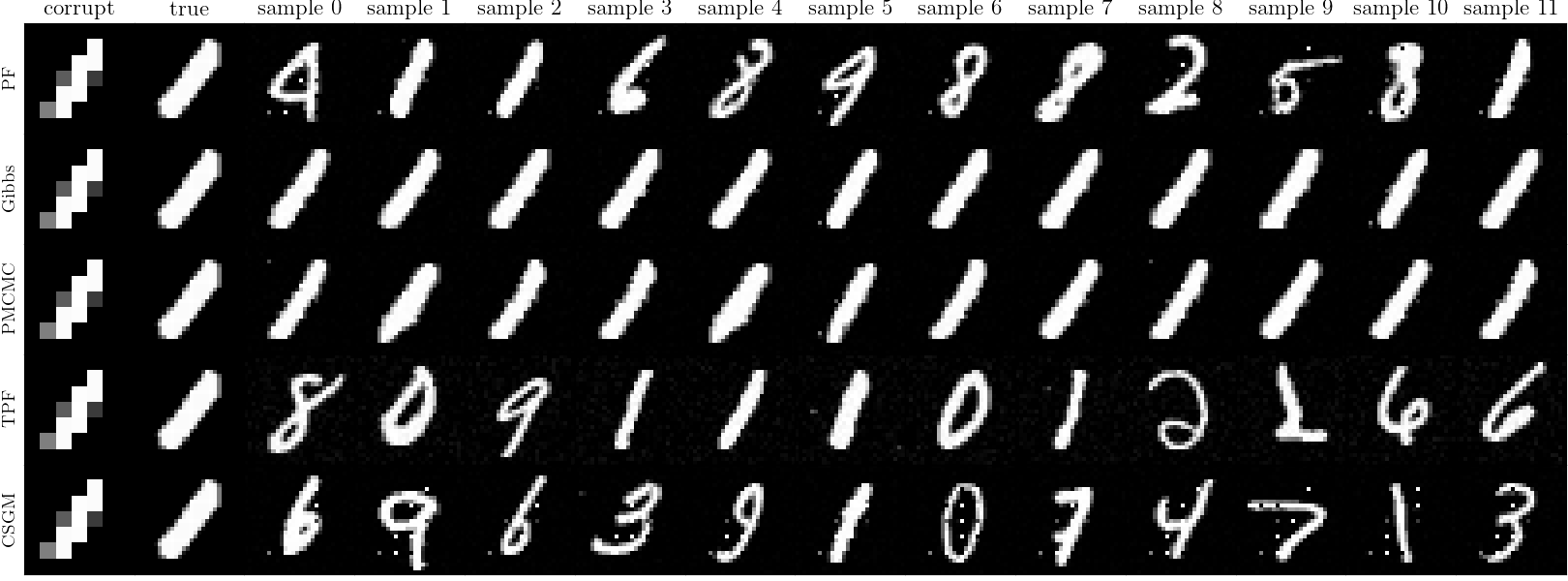}
    \includegraphics[width=.8\linewidth]{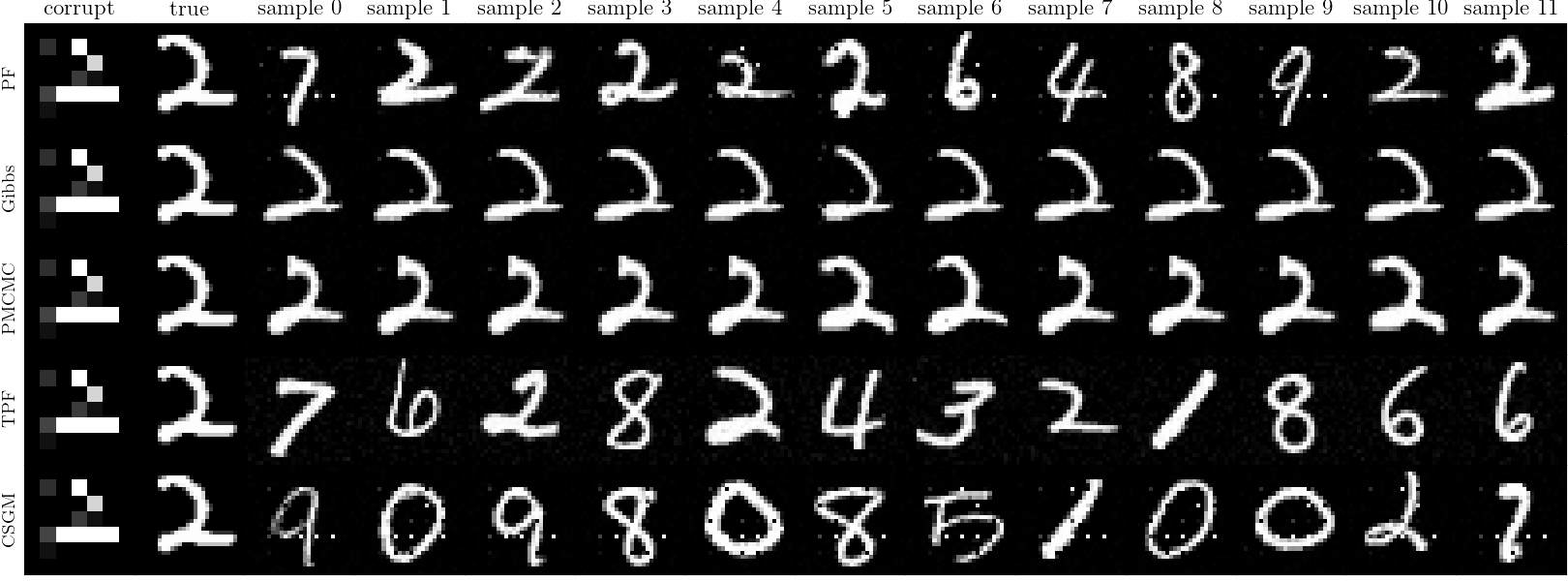}
    \includegraphics[width=.8\linewidth]{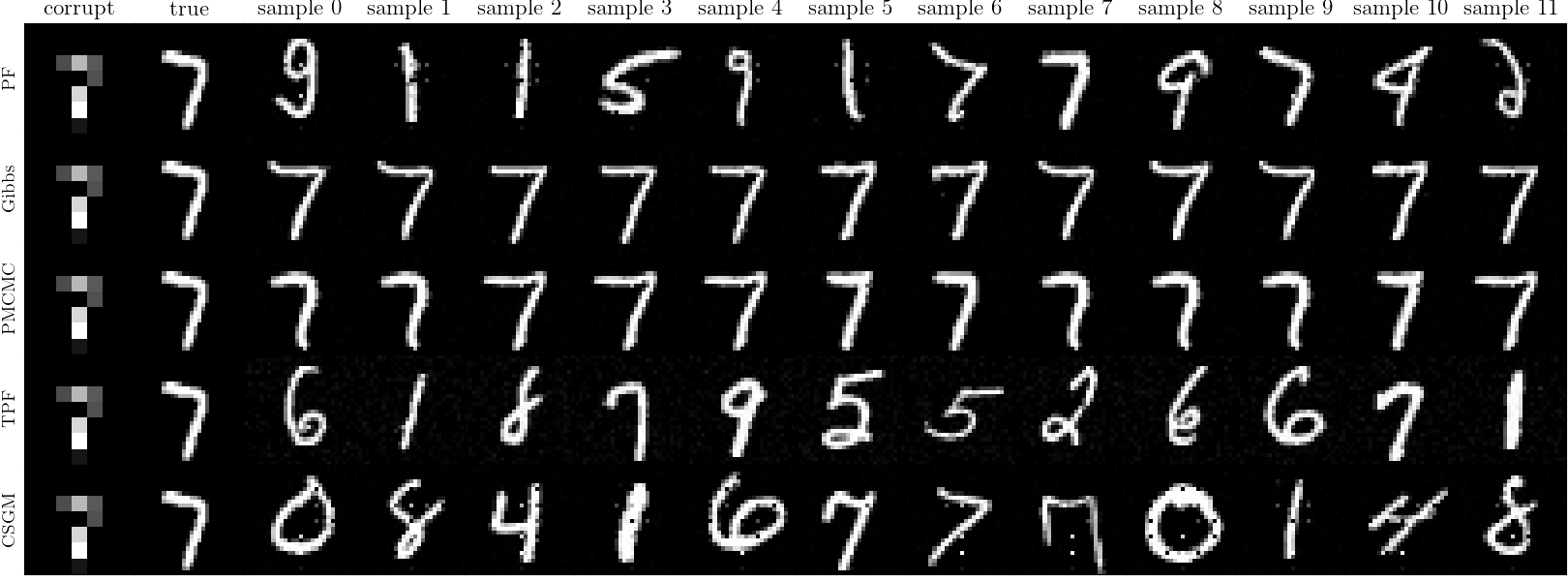}
    \includegraphics[width=.8\linewidth]{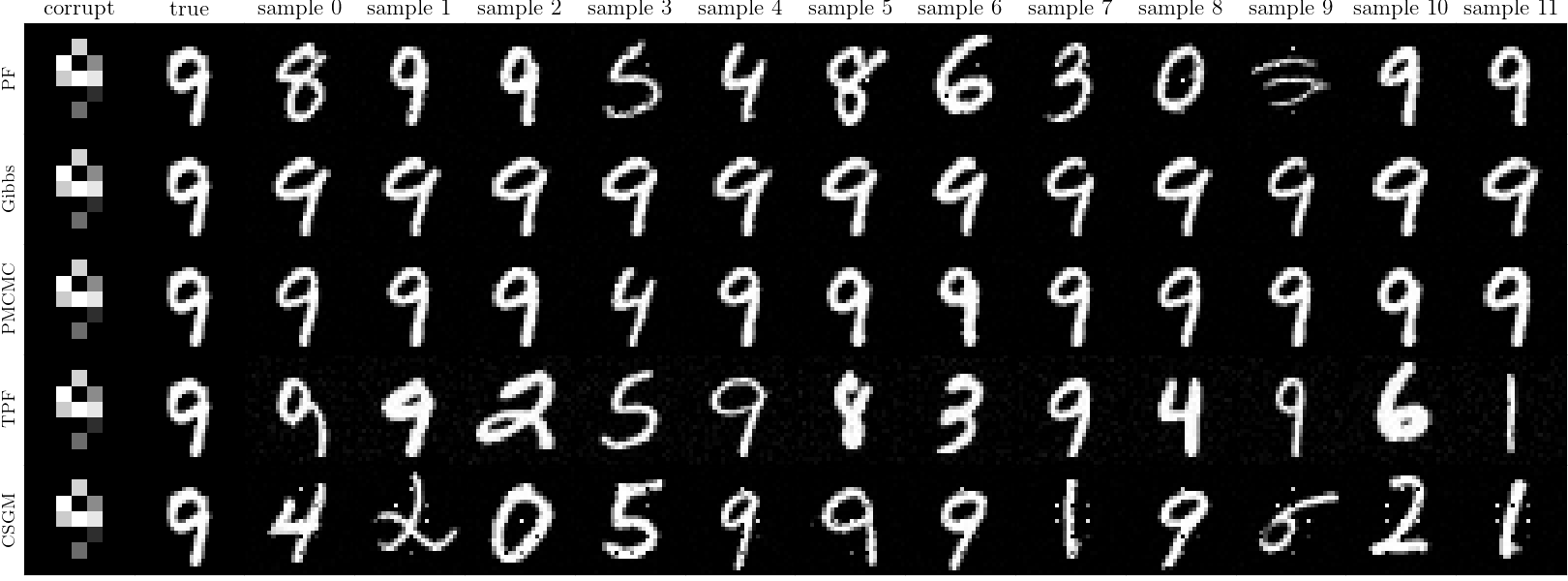}
    \caption{Examples of MNIST super-resolution tasks. The figure has four panels, showing the results for four test images. For each panel, the first to last rows show the results of PF, Gibbs-CSMC, PMCMC-0.005, TPF, and CSGM, respectively. Wrong samples and artefacts appear less in our MCMC methods compared to others.}
    \label{fig:app-mnist-supr}
\end{figure}

\begin{figure}
    \centering
    \includegraphics[width=.8\linewidth]{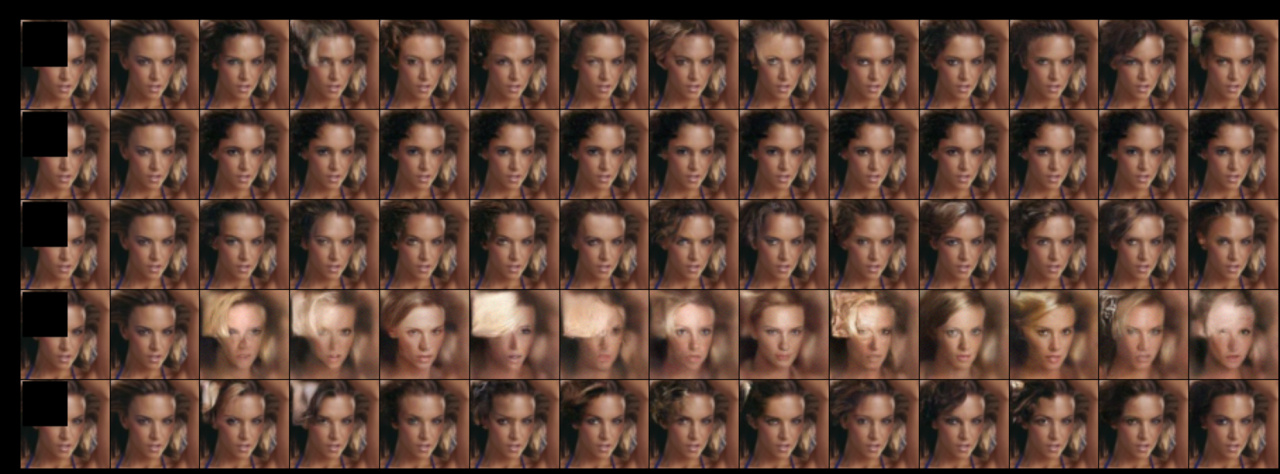}
    \includegraphics[width=.8\linewidth]{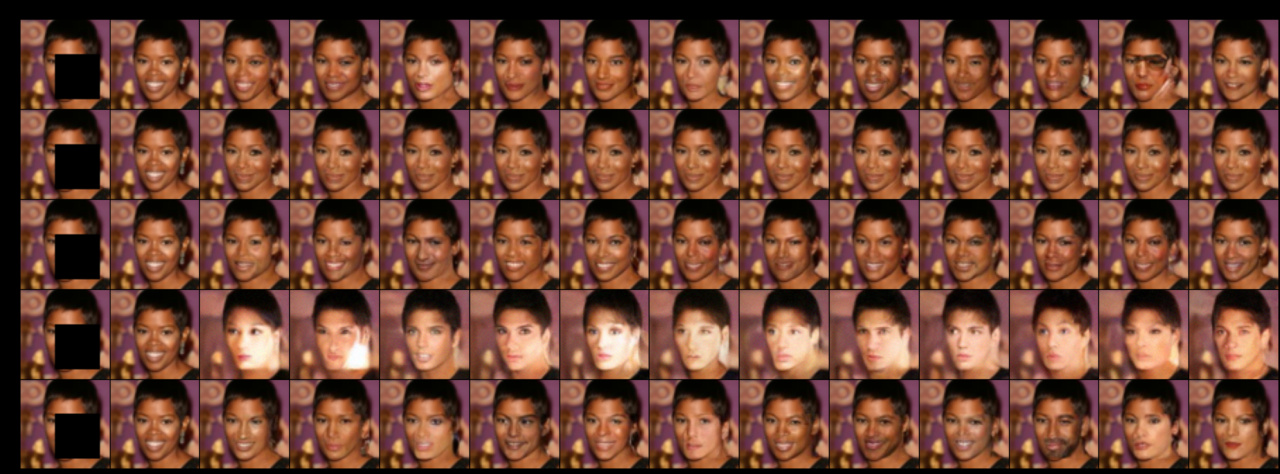}
    \includegraphics[width=.8\linewidth]{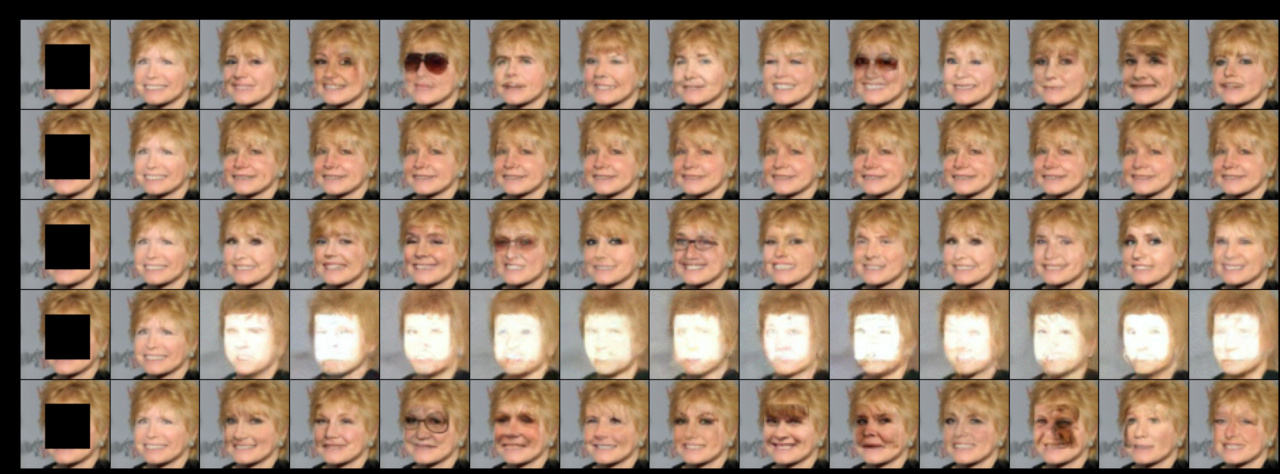}
    \includegraphics[width=.8\linewidth]{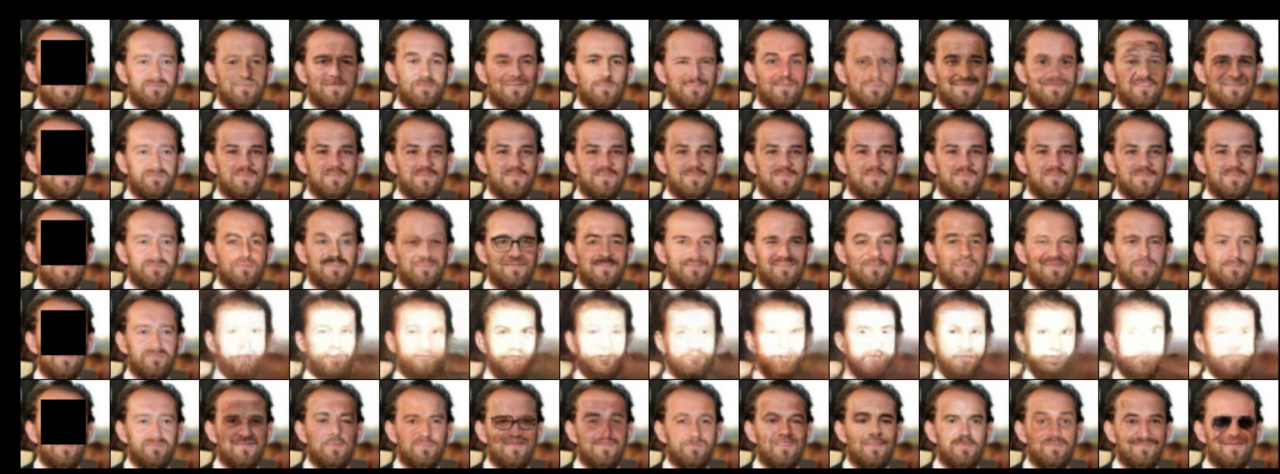}
    \caption{Examples of CelebaHQ inpainting tasks. The figure has four panels, showing the results for four test images. For each panel, the first to the last rows show the results of PF, Gibbs-CSMC, PMCMC-0.005, TPF, and CSGM, respectively. }
    \label{fig:app-celeba-inpainting}
\end{figure}

\begin{figure}
    \centering
    \includegraphics[width=.8\linewidth]{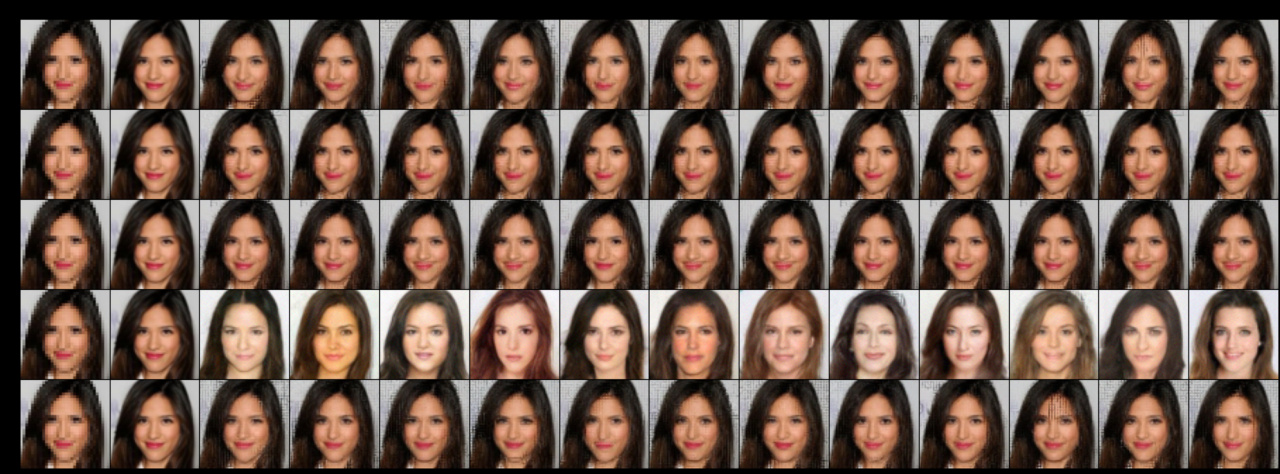}
    \includegraphics[width=.8\linewidth]{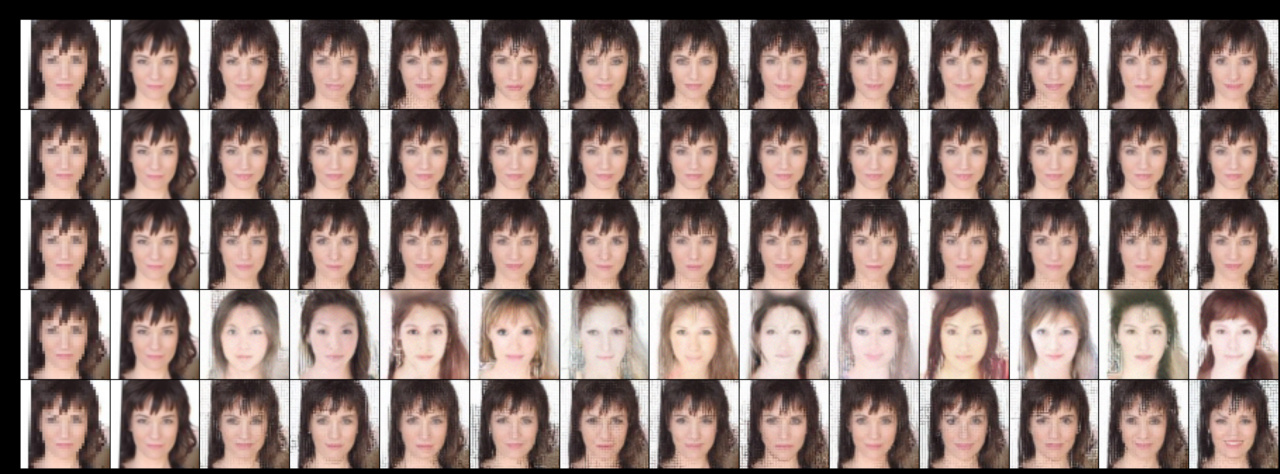}
    \includegraphics[width=.8\linewidth]{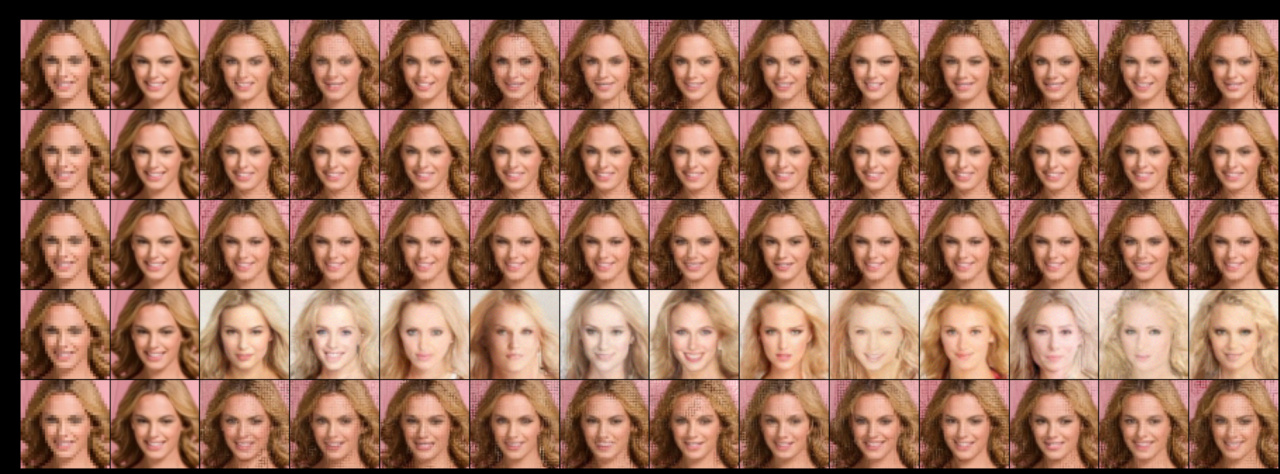}
    \includegraphics[width=.8\linewidth]{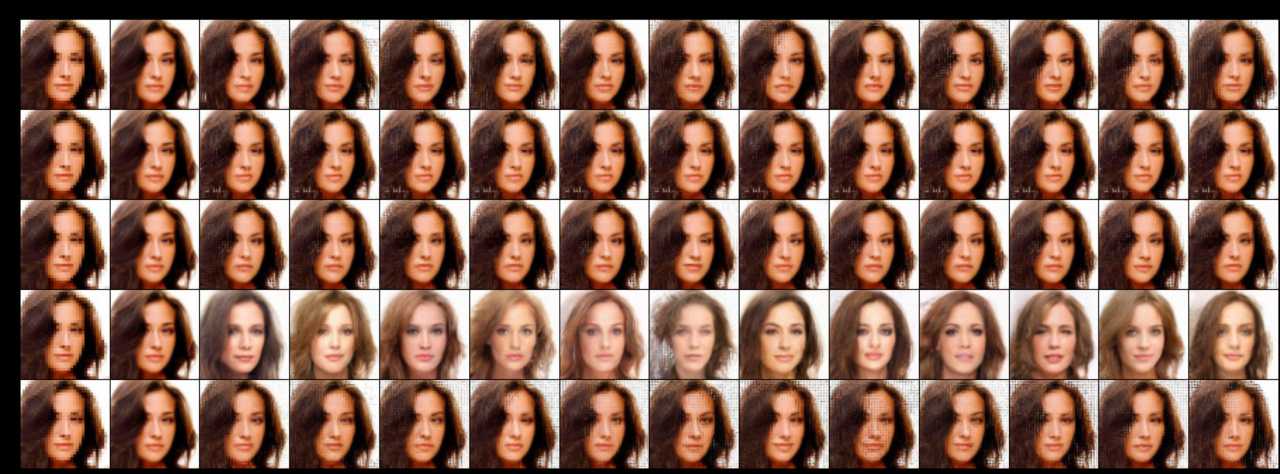}
    \caption{Examples of CelebaHQ super-resolution tasks. The figure has four panels, showing the results for four test images. For each panel, the first to the last rows show the results of PF, Gibbs-CSMC, PMCMC-0.005, TPF, and CSGM, respectively. }
    \label{fig:app-celeba-supr}
\end{figure}

We show samples of inpainting and super-resolution tasks in Figures~\ref{fig:app-mnist-inpainting}--\ref{fig:app-celeba-supr}, to qualitatively compare the methods. 
We find from these figures that our Gibbs-CSMC and PMCMC methods indeed outperform all other methods, in terms of quality and correctness.\footnote{While there are qualitative scoring methods~\citep[e.g.,][]{wang2023exploring, Ke2021ICCV}, we here focus on reference-based quantitative comparisons.} 
By correctness, we mean that the samples generated by our methods are realistic and closer to the true images. 
As an example, in the first panel of Figure~\ref{fig:app-mnist-inpainting}, the MCMC chains generate a variety of digits ``4'' which indeed correspond to the true digit, while other methods can generate wrong, albeit more diverse digits.

\begin{figure}[t!]
    \centering
    \includegraphics[width=.99\linewidth]{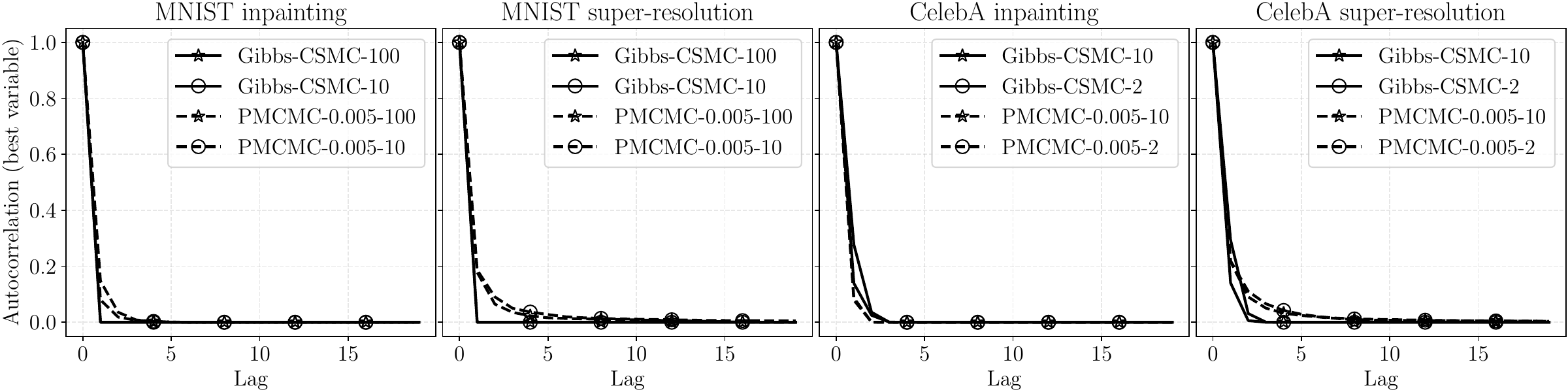}
    \caption{Autocorrelations of Gibbs-CSMC and PMCMC (with $\delta=0.005$) for image inpainting and super-resolution. The autorrelations are averaged over 100 individual test samples. For each test sample, its autocorrelation is computed over 100 MCMC samples, and we select the best curve among all the image pixels. We select the best curve because some pixels will not change at all given the problem definition. 
    The autocorrelation of Gibbs-CSMC is marginally better than PMCMC. }
    \label{fig:app-autocorr-imgs}
\end{figure}

The autocorrelations of Gibbs-CSMC and PMCMC methods are shown in Figure~\ref{fig:app-autocorr-imgs}. 
We see that both MCMC methods admit reasonable autocorrelations in all the tasks, while Gibbs-CSMC is better than PMCMC. 

\section{Particle filtering coalescence}
\label{sec:app-coalescence}
\begin{figure*}[t!]
    \centering
    \includegraphics[width=.99\linewidth]{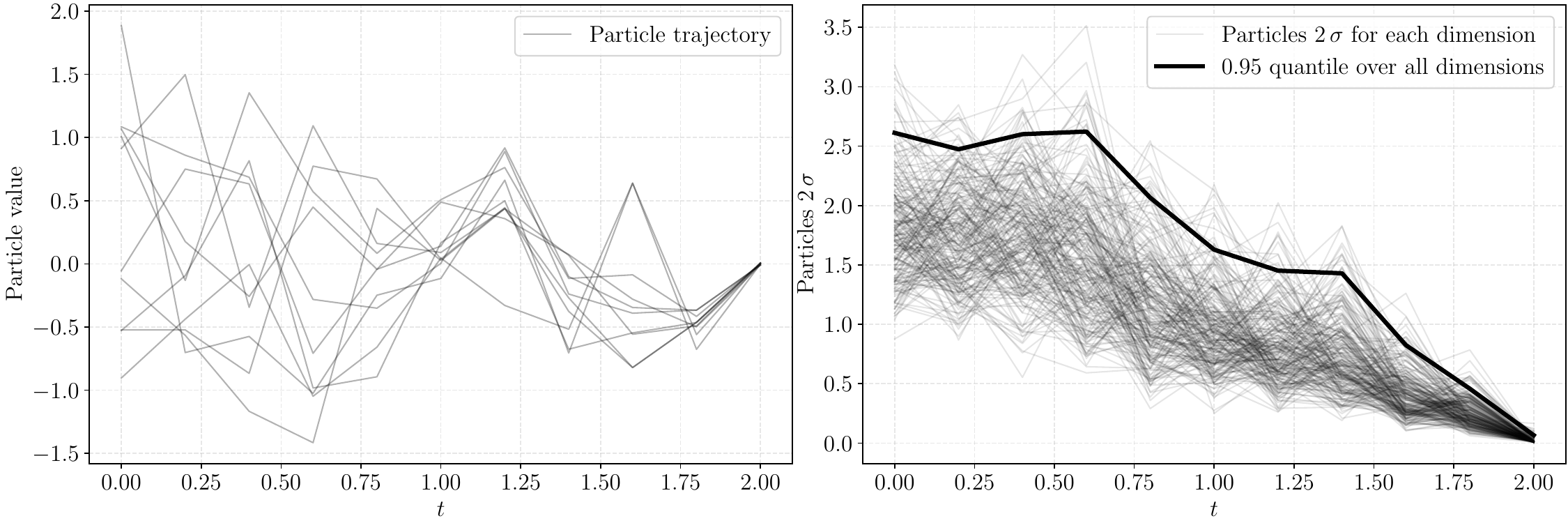}
    \caption{The coalescence effect of particle filtering (with 10 particles) on MNIST inpainting. 
    The left figure shows the trajectories of the 10 particles in the filtering pass, while the right figure shows the two standard deviation of all particles for each dimension. 
    We see that all the particles collapse to one single point at the terminal time, and this holds true across all the state dimensions. }
    \label{fig:app-coalescence}
\end{figure*}

Lastly, we would also like to discuss a coalescence effect of the particle filter approach for conditional sampling~\citep{trippe2023diffusion, dou2024diffusion}. 
This effect is reflected in Figure~\ref{fig:app-coalescence}, where we see that all particles collapse to a single point at the terminal time of the filtering pass. 
Consequently, this means that it is hard to treat all the particles at the terminal time as ``independent'' samples (after resampling). 
More precisely, to draw $J$ independent samples, we would have to run the PF $J$ times independently, instead of running the PF once and using all the terminal particles as approximately the independent samples. 

\end{document}